\documentclass[11pt,a4paper]{article}
\usepackage[a4paper,margin=25mm]{geometry}
\usepackage[T1]{fontenc}
\usepackage[utf8]{inputenc}
\usepackage{newtxtext,newtxmath}
\usepackage{amsmath,graphicx,booktabs,longtable,array,setspace,caption,needspace,lineno,float,placeins}
\usepackage[super,comma,sort&compress]{natbib}
\usepackage{xurl}
\usepackage[unicode,hidelinks]{hyperref}
\raggedright
\hypersetup{pdftitle={Oscillatory Processing Units as a computing substrate for biomimetic intelligence},pdfauthor={Alexander A. Wolf},pdfsubject={Research preprint}}
\begin{document}
{\Large\bfseries Oscillatory Processing Units as a computing substrate for biomimetic intelligence\par}
\vspace{10pt}
\textbf{Alexander A. Wolf*}\\
Lead Researcher, Artificial Brains\\
San Francisco, United States\\
*Correspondence: \href{mailto:research@artificialbrains.ai}{research@artificialbrains.ai}\\
ORCID: \href{https://orcid.org/0009-0005-4975-289X}{https://orcid.org/0009-0005-4975-289X}
\vspace{6pt}

\section*{Abstract}
Biological intelligence emerges from neural activity that connects sensation, memory and action. Synaptic plasticity, neuronal assemblies and oscillations provide principles for biomimetic computing, but translating them into reusable operations requires a defined computational unit. We introduce Oscillatory Processing Units (OPUs), groups of synthetic neurons sharing a rhythm. Their directed interactions transform incoming activity and prior receiving state into an Oscillatory Occupancy Space (OOS), which becomes available to subsequent computation. We show that source--receiver conditions and oscillatory organization shape these transformations, including their geometry, persistence and historical accessibility. An Atlas of measured relationships supports prospective composition with predicted response classes. Configured circuits perform logical and numerical operations, express associative recall and generate closed-loop decisions with fixed synaptic weights. OPUs and OOS establish a computing substrate for biomimetic intelligence in which experience remains part of the state through which new input becomes action.

\section*{Introduction}
Biological brains connect sensation, memory and action through the activity of interacting neurons. Neuroscience has identified several mechanisms that contribute to this computation. Activity-dependent synaptic modification provides a physiological basis for Hebbian learning.\cite{bi1998} Engram experiments link the reactivation of specific neuronal assemblies to remembered behavior.\cite{liu2012} Intrinsic cellular oscillations provide local temporal structure,\cite{alonso1989} and phase-dependent population activity organizes information about remembered events.\cite{siegel2009} Rhythmic synchronization between neuronal groups modulates how effectively one group drives another,\cite{fries2015} and nested theta and gamma oscillations can organize multiple items in temporal order.\cite{lisman2013} Recurrent population dynamics support context-dependent decisions,\cite{mante2013} while hidden neural states can preserve information for later memory-guided behavior.\cite{wolff2017} Together, these findings connect intelligence to changing synaptic relationships, collective state and the temporal organization of activity. Translating them into a computing substrate requires a unit that specifies how an incoming signal acts on an existing state and what becomes available to the next operation. Such a unit must support both measurement and construction: its transformation should be characterizable, its operating conditions identifiable, and its output usable by another unit. The problem addressed here is to realize such a unit and establish what can be built from it.

Dynamical computing provides important precedents. Liquid-state machines make present and past inputs available through transient recurrent activity,\cite{maass2002} and echo-state networks use recurrent dynamics with trained output mappings.\cite{jaeger2004} Coupled recurrent networks implement persistent states and input-driven transitions.\cite{rutishauser2009} The Neural Engineering Framework connects representation, transformation and dynamics in neural populations.\cite{eliasmith2003} The Assembly Calculus formalizes brain computation as operations on neuronal assemblies, such as projection, association and merge, realized through Hebbian plasticity and inhibition.\cite{papadimitriou2020} Oscillatory neurocomputers use frequency relationships and input-dependent coupling for associative computation.\cite{hoppensteadt1999} Physical oscillator devices have performed spoken-digit recognition,\cite{torrejon2017} and networks of coupled spin-torque oscillators have recognized spoken vowels through trained synchronization.\cite{romera2018} Reviews of physical reservoir computing and oscillator-based computing survey the breadth of these hardware approaches.\cite{tanaka2019,csaba2020} Biomimetic micro-assemblies have been composed into circuits supporting memory, learning and decisions.\cite{pathak2026} Loihi connects a specified spiking computational model to a programmable architecture and evaluates that architecture through implemented computations.\cite{davies2018} These approaches establish ways to compute with neural and physical dynamics. Our focus is the directed interaction between populations: characterizing its transformation under specified conditions, then using that characterization to select an interaction for a new circuit before the circuit is run.

We propose Oscillatory Processing Units (OPUs), populations of synthetic neurons sharing a rhythm, as a computing substrate for biomimetic intelligence. The central proposition is that oscillatory organization helps define the transformation one OPU performs on the state of another, the Oscillatory Occupancy Space (OOS), allowing measured interactions to serve as building blocks for artificial brains.

We examine this proposition through characterization, intervention and construction. The factorial Atlas maps how source and receiver conditions shape the OOS. Temporal interventions test the role of oscillatory organization, and prospective composition tests whether measured relationships predict the response classes of a new circuit. Memory experiments examine persistence, historical accessibility and associative response. Controlled collision avoidance and a physical ANT robot connect these operations to action. Biology supplies the organizing principles; the empirical contribution is a realized computational framework that links state transformation, memory and behavior. Its broader implication is a route toward intelligence built through continuing interaction with experience.

\section*{Computational model}
We introduce the \textbf{Oscillatory Processing Unit (OPU)}, a group of synthetic neurons sharing a rhythm. The directed interaction of a source OPU and a target OPU is the \textbf{computational unit}. Incoming activity acts on the receiver's preceding state, generating an \textbf{Oscillatory Occupancy Space (OOS)}: the evolving activity, recurrence and reachable state transitions of the receiving population. That state becomes available to subsequent interactions (Figure 1).

A \textbf{synthetic neuron} is an engineered spiking element with evolving internal state and intrinsic temporal dynamics. It receives events through directed connections and emits output events. The synthetic-neuron model is proprietary, but its intrinsic dynamics can be described functionally. Each neuron combines synaptic input accumulation, internal state evolution, and oscillatory modulation of its response. Input can influence the neuron beyond the instant at which it arrives, while accumulated influence decays over time. The internal oscillatory cycle modulates how this influence contributes to subsequent output, making the neuron’s response dependent on its current phase and recent input history. Consequently, identical inputs arriving at different points in the cycle, or following different preceding inputs, can produce different responses. This temporal sensitivity is intrinsic to the neuron and does not require synaptic weights to change. We use these elements to construct OPUs and measure their collective response through the emitted spike record.

Source and target can belong to the same or different rhythm classes, and neurons within an OPU can express distinct activity patterns. The nominal cycle periods in the factorial and native-lag experiments are 5 model steps for Gamma, 15 for Beta, 22 for Alpha, 50 for Theta and 150 for Delta. The rhythm names label these configured periods and do not refer to biological frequency bands. These configured periods supply the cycle references used in the temporal analyses. The stimulus record specifies the delivered input events. The reported fixed-weight runs hold synaptic weights constant while neural and short-term synaptic state evolve.

TNorm is a dimensionless measure of synaptic strength normalized to the operating range of a directed source--target OPU interaction. It specifies the relative operating position of that interaction, from the predicted onset of assembly contact (TNorm $=0$) to a calibrated upper reference associated with assembly compression and fusion (TNorm $=10$). This normalization enables comparison across OPU interactions with different rhythms and connection properties, without implying equivalent neural activity or computation. When a TNorm value is assigned to an OPU, it refers to that OPU’s incoming interaction with a specified source or reference input. We use TNorm as the operating coordinate of each interaction: the Atlas sweeps it across this range to map how a transformation changes with operating position, and new circuits are specified by selecting TNorm values from those maps.

The recorded output of an OPU is a sequence of population spike patterns. At each model step, the identities of the neurons that fire define the active assembly. We characterize the OOS through the assemblies visited, their occupation frequencies, their transitions and their relationships across time. These observations expose the geometry, persistence and historical accessibility of the receiving state.

During execution, incoming events act on the receiver's current state. The resulting activity drives its outgoing connections and contributes to subsequent states. When a target OPU becomes the source of another interaction, its activity carries the effects of the preceding pathway into the next operation. A configured circuit therefore couples its operations through directed connections and continuing neural state.

The relation is specified by its source, receiver, temporal classes, operating points, directed connection and preceding state and input history. We evaluate it through the transformation of receiving state and what a downstream circuit can use, consistent with measuring information-processing capacity in a dynamical system.\cite{dambre2012} A target can become the source of another relation, allowing operations to be composed while retaining history in the intervening state. Changing the immediate source, feedback or initial state changes the conditions under which the operation is evaluated.

Figure 1 expresses the population-level relationship as $\mathcal F_{A\rightarrow B}$. Source and receiving states are denoted by $\widehat{\mathcal O}_A$ and $\widehat{\mathcal O}_B$; $r_A,r_B$ identify rhythm classes, $u_A,u_B$ their operating points and $M_{AB}$ their connection. The symbols $N,n$ denote population sizes and $w$ denotes connection strength. The map describes the transformation under the stated conditions and activity history.

\begin{figure}[H]
\centering
\includegraphics[width=\linewidth,height=.55\textheight,keepaspectratio]{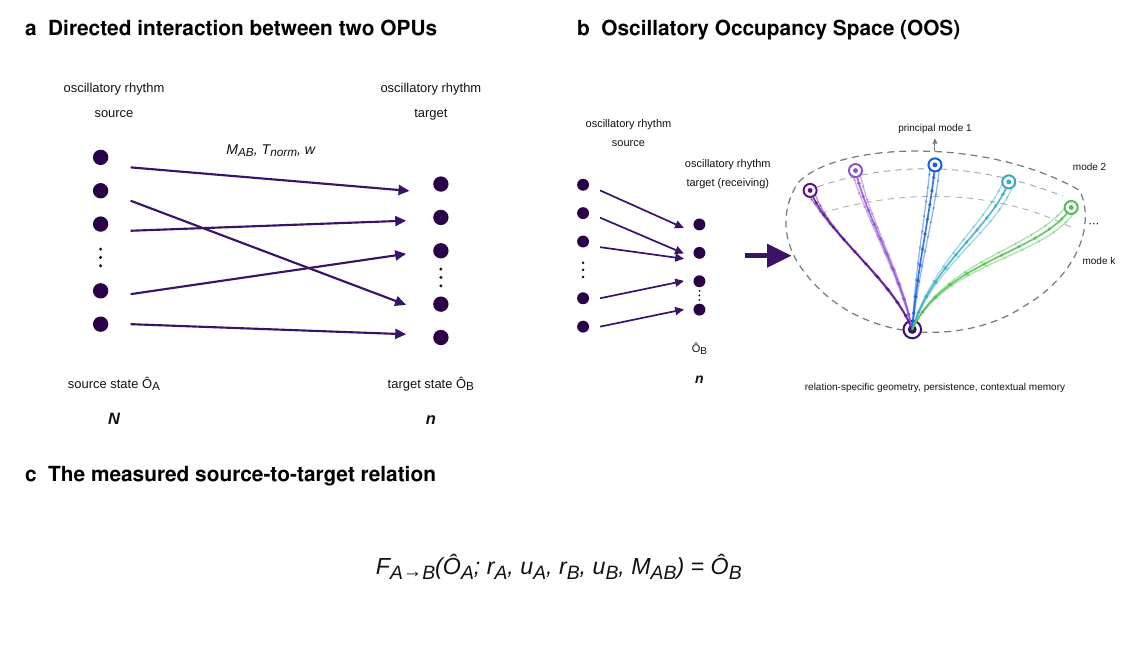}
\begin{spacing}{1.0}
\caption{\textbf{OPU interactions generate an evolving OOS.} \textbf{a,} A source OPU projects to a receiving OPU. $N$ and $n$ denote population sizes, $M_{AB}$ the directed connection, $w$ its strength and TNorm its normalized operating coordinate. \textbf{b,} Schematic OOS geometry illustrates relation-specific state organization, persistence and contextual accessibility. \textbf{c,} The population-level mapping describes the source--receiver transformation under their rhythm classes, operating points, connection and preceding activity history. All panels are conceptual schematics.}
\label{fig:1}
\end{spacing}
\end{figure}
\FloatBarrier

\section*{Results}

\subsection*{Observable population responses}
We characterized the substrate through controlled stimulus--response measurements at the population and directed-interaction levels. First-level populations were observed under a fixed sensory schedule while their operating coordinates were swept over fixed connectivity. Recruitment curves report rhythm-normalized participation across each population's operating range. This readout scales cycle-averaged active-neuron fraction by the configured release fraction (Methods, Population-state measurements). The same event records supply assembly diversity, effective dimensionality and recurrence (Figure 2c--e). Together, these measurements characterize the amount and organization of the response.

Directed-pair measurements describe how a source shapes the receiving OOS: alignment with source activity, information shared between states, persistence across lags and accessibility of earlier input. Paired temporal interventions establish how those responses depend on oscillatory organization. Table 1 connects the measured properties to the experiments below. These measurements provide an empirical basis for selecting and composing OPU interactions.

\begin{table}[H]
\caption*{\textbf{Table 1. Observable measures of OPU computation.} Each property is defined through the recorded response of a configured population or directed interaction. The cited experiments supply its measured characterization.}
\small
\begin{tabular}{@{}p{0.24\linewidth}p{0.48\linewidth}p{0.21\linewidth}@{}}\toprule
\textbf{Property} & \textbf{Measurement} & \textbf{Evidence}\\\midrule
Population operating response & Participation across operating coordinates under a fixed input schedule and connectivity. & Figure 3a\\[5pt]
State organization & Assembly diversity, effective rank and recurrence of the receiving population. & Figures 2 and 3b\\[5pt]
Directed transformation & Source--target alignment, information, fusion and fragmentation. & Figures 2 and 3c\\[5pt]
Composition & Response classes of new circuits predicted from measured relations. & Figure 4\\[5pt]
Historical dependence & State persistence, earlier-input accessibility and terminal associative responses. & Figures 5 and 6\\[5pt]
Action & Neural decisions and closed-loop behavior. & Figure 7; Supplementary Video 1\\[5pt]
\bottomrule
\end{tabular}
\end{table}
\FloatBarrier

\subsection*{Source--receiver conditions explain state beyond activity}
We tested whether source and target operating conditions explained receiving-state organization beyond firing participation. All 16 directed pairings of four rhythm classes were evaluated on an 11-by-11 grid across 11 independently generated networks. Participation, state geometry, recurrence and source--target relationships were measured separately.

Across 240 pair-by-descriptor comparisons, mean conservative held-out $R^2$ increased from 0.487 for source and target participation alone to 0.724 when their operating coordinates were also retained. Source--receiver conditions therefore explained additional structure in the receiving state. Descriptors with poor fitted prediction were retained as empirical measurements (Methods, Factorial characterization and Atlas; Supplementary Figure 1).

The fifth-rhythm extension completed 25 directed pairings, each measured at 121 operating points across the same 11 networks, yielding 3,025 operating cells. Figure 2 shows how source alignment, target dimensionality and recurrent organization vary between directed relationships and across the operating surface of each relationship. These measurements form the neural-dynamics Atlas: a map of OPU transformations under specified conditions.

\begin{figure}[H]
\centering
\includegraphics[width=\linewidth,height=.67\textheight,keepaspectratio]{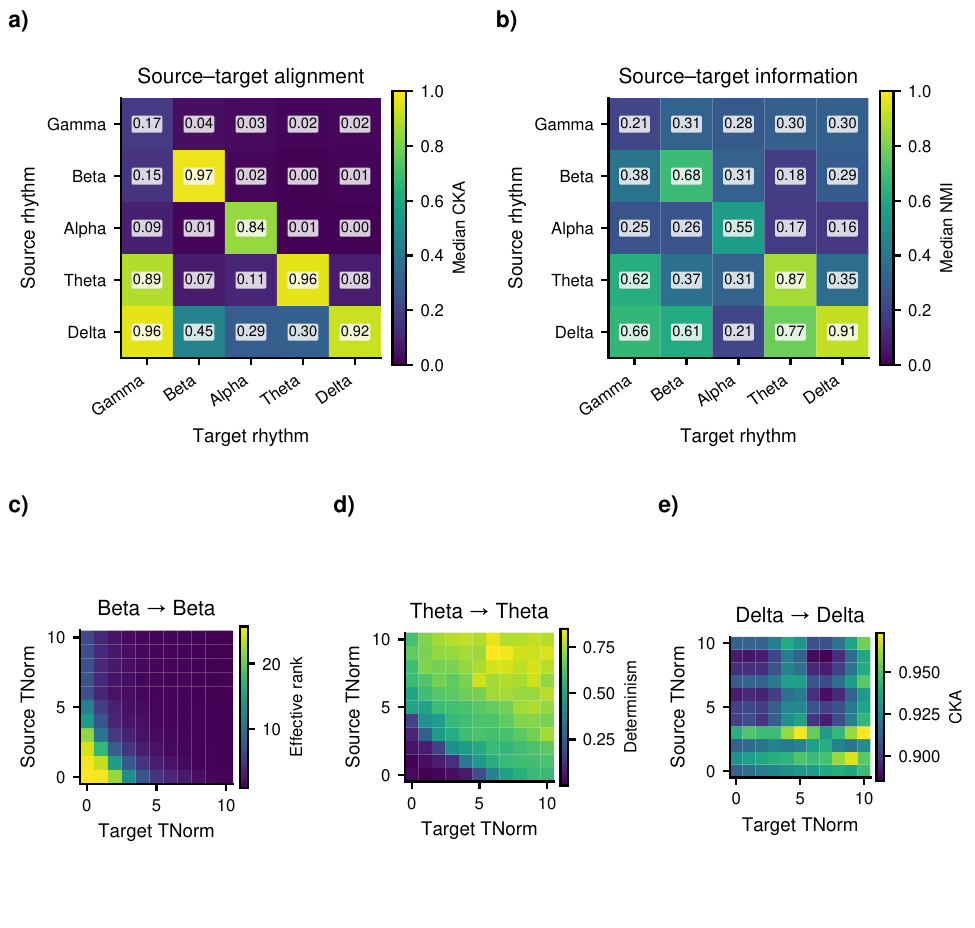}
\begin{spacing}{1.0}
\caption{\textbf{The neural-dynamics Atlas maps directed OPU transformations.} \textbf{a,b,} Median source--target CKA and NMI across the 121 operating points in each of the 25 directed rhythm pairs, using the Gamma-mediated input path. Rows denote source rhythm and columns denote receiving rhythm. \textbf{c--e,} Measured 11-by-11 operating maps for Beta-to-Beta effective rank, Theta-to-Theta determinism and Delta-to-Delta CKA. Each coordinate contains its summary across 11 independently generated networks. The maps display measured cells; pair-level panels summarize those coordinates by their medians. CKA and NMI quantify representational alignment and normalized mutual information. The four-class predictive-model comparison appears in Results and Supplementary Figure 1. Selected cells used in the prospective composition are listed in Table 2.}
\label{fig:2}
\end{spacing}
\end{figure}
\FloatBarrier

A second five-rhythm factorial changed the input pathway while preserving sensory observations. The resulting state depended on the activity reaching the source. Matching pathways by realized source and target participation improved held-out source--target alignment prediction from $R^2=0.909$ to 0.939 across 3,025 cells (Supplementary Figure 2). This transfer between input-path architectures establishes the importance of the incoming state when reusing an operating coordinate.

The Atlas links these response measurements to computational uses: preserving distinctions, combining signals, maintaining context and changing downstream timing. Complete circuits establish the logical, associative and motor operations that selected interactions support.

\subsection*{Oscillatory organization changes the transformation}
We tested the contribution of phase-dependent sensitivity and periodic release to OPU computation. Phase averaging replaced within-cycle variation in neuronal sensitivity by its cycle mean. Release averaging replaced the periodic control of when retained input contributes to the response by its cycle mean. The paired study compared intact timing, phase averaging, release averaging and their combination, with populations, structural connections, synaptic weights, sensory schedule and remaining state dynamics retained (Methods, Oscillatory organization and activity-matched comparisons).

Substantial recruitment survived the intervention. Populations continued to respond across the operating sweep. State organization changed: in the illustrated Beta condition, entropy increased from 6.53 to 11.18 bits and effective rank from 1.97 to 10.74, replacing a compact response with a distributed one. Theta and Delta also changed their dimensionality and recurrence (Figure 3a,b).

Directed relations changed differently. Delta-to-Delta and Theta-to-Theta lost source--target alignment and information. In Delta-to-Theta, alignment increased from 0.309 to 0.791 while normalized mutual information fell from 0.767 to 0.388 (Figure 3c). Temporal averaging produced a more directly aligned, less differentiated transformation in this relation.

These relational differences persisted in a secondary analysis matching source and target participation within the same relation and network. The comparison controls mean recruitment while leaving temporal activity and effective input free to vary. Persistence also fell: after one target cycle, null-centered receiving-state persistence changed from 0.740 to 0.137 in Delta-to-Delta, 0.577 to 0.091 in Delta-to-Theta and 0.414 to 0.044 in Theta-to-Theta (Figure 3d).

A complementary frozen-profile analysis found loss of reference-profile support in all nine specified contrasts across three source--Theta relations, with the same direction in all 11 networks and no classifier refitting (Supplementary Figure 3). Both analyses link temporal organization to the geometry and persistence of the receiving state.

Connection strength also positioned the interaction within different regimes, including recruitment, compression, competition, suppression and saturation. Temporal organization and connection conditions jointly specify the operation performed by an OPU interaction.

\begin{figure}[H]
\centering
\includegraphics[width=\linewidth,height=.68\textheight,keepaspectratio]{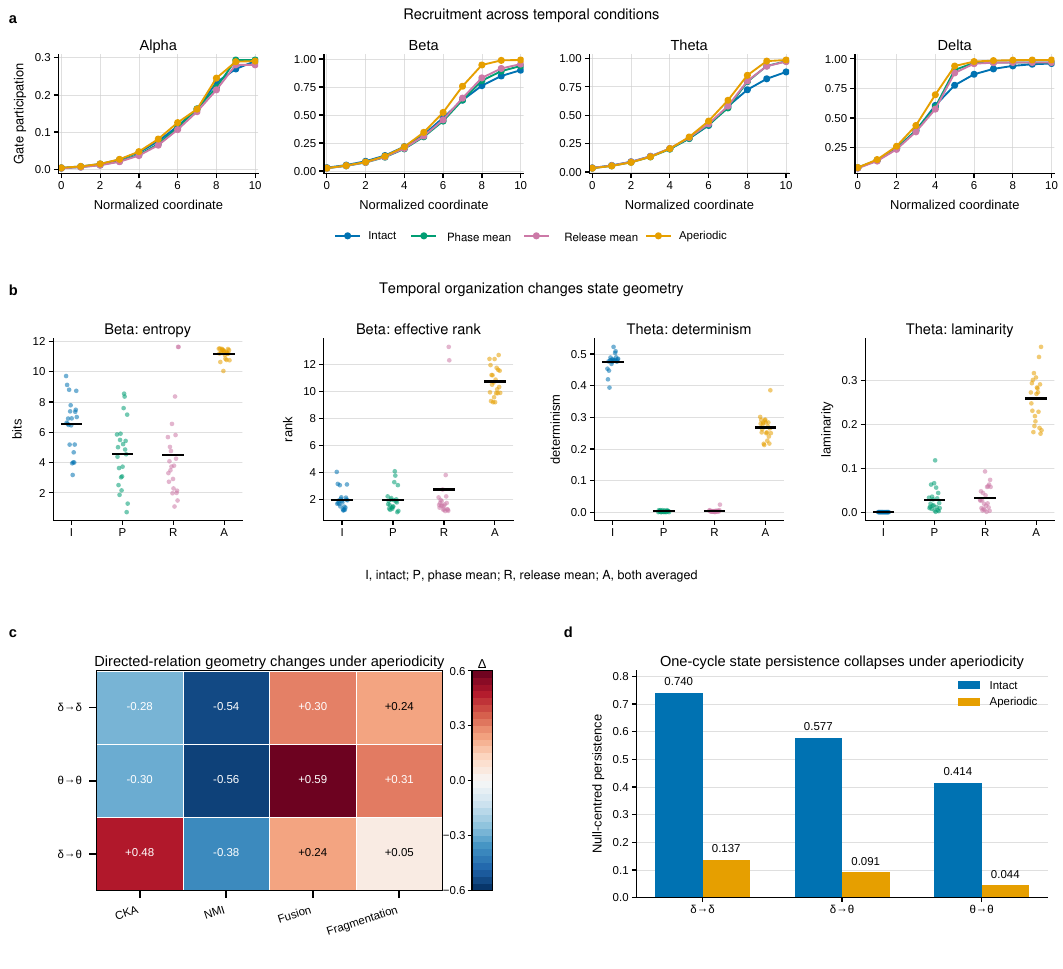}
\begin{spacing}{1.0}
\caption{\textbf{Temporal interventions reshape population state across sustained recruitment.} \textbf{a,} Rhythm-normalized participation across the operating coordinate for Alpha, Beta, Theta and Delta under intact timing, phase averaging, release averaging and both averaged. The key is I, intact; P, phase mean; R, release mean; A, both averaged. The normalization is defined in Methods. \textbf{b,} Paired observations and group means for representative Beta entropy and effective rank and Theta determinism and laminarity, with operating settings held fixed within each comparison. \textbf{c,} Altered-minus-intact changes in source--target alignment (CKA), normalized mutual information (NMI), fusion and fragmentation for three directed relations. \textbf{d,} Null-centered persistence after one target cycle: Delta-to-Delta changes from 0.740 to 0.137, Delta-to-Theta from 0.577 to 0.091, and Theta-to-Theta from 0.414 to 0.044. The paired temporal-ablation study used 11 networks. Recruitment matching is a secondary analysis described in Methods. ``Aperiodic'' denotes both features averaged in the manipulated populations; the remaining components retain their temporal dynamics.}
\label{fig:3}
\end{spacing}
\end{figure}
\FloatBarrier

\subsection*{Configured interactions perform explicit operations}
Selected OPU interactions performed identifiable transformations of independently supplied operands. In 152 technically valid runs across eight networks, bounded addition generalized across the tested sensory identities: mean leave-one-identity-out $R^2$ was 0.953 and 0.892 for two constructions, compared with fitted values of 0.954 and 0.903. These regressions evaluated the operand--output relationships produced by the configured neural circuits.

OR, AND and VETO met the strict separation criterion in every tested identity--network combination for one substrate. Signed subtraction achieved $R^2=0.940$ and 93.4\% sign agreement. The tested XOR construction met its criterion in 45 of 72 identity cases, with the principal shortfall in both-active inputs. Unequal effective source amplitudes were identified as a candidate explanation for subsequent testing (Methods, Operations and fuzzy logic).

Separate engineering panels examined graded and decision outputs. Fuzzy AND, OR and NOT reproduced the expected patterns at five tested input levels across three networks, including one downstream reuse. A shallow decision panel preserved tested rule decisions, support ordering and ties at its neural outputs. A selected composition linking recent context, competition and action commitment produced the intended history-conditioned mapping in eight of eight fresh sessions, against five of eight in a zero-history-coupling sham. Methods, Decision-layer studies, describes the individual evaluations.

The operations provide ways to connect evidence to subsequent computation: a numerical comparison can supply a graded condition, that condition can support an action, and a veto can suppress an incompatible alternative. Their scope is defined by the tested inputs and constructions. The stronger test of a reusable unit is whether measurements select a new composition before its response is observed.

\subsection*{Measured relations specify a new composition}
We selected a three-stage circuit from the pairwise Atlas, froze its operating coordinates and response-class predictions, and evaluated 11 new networks under a held-out sensory schedule. A Theta source at TNorm 10 drove a shared Delta intermediate at TNorm 9, which drove Alpha, Beta and Gamma receivers at TNorm 1, 10 and 6, respectively. Isolated Delta-to-terminal references and direct Theta-to-terminal skip paths provided controls (Figure 4a). No coordinate sweep, retraining or post-observation retuning was performed.

The selected Atlas cells had source--target CKA values of 0.620 for Delta-to-Alpha, 0.056 for Delta-to-Beta and 0.967 for Delta-to-Gamma (Table 2). At these coordinates, Alpha therefore had greater alignment than Beta. Figure 2a summarizes the complete relation surfaces: their median CKA values were 0.293 and 0.446, respectively. The local operating conditions determine the prospective ordering; the surface medians describe the broader range of each relation.

\begin{table}[H]
\caption*{\textbf{Table 2. Selected Atlas cells and prospective outcomes.} The shared intermediate was Theta(10)-to-Delta(9). Atlas values are the Gamma-mediated summaries at the frozen coordinates across 11 reference networks. Prospective values are means from the composed circuit across 11 new networks.}
\small
\begin{tabular}{@{}p{0.24\linewidth}p{0.16\linewidth}p{0.16\linewidth}p{0.15\linewidth}p{0.15\linewidth}@{}}\toprule
\textbf{Directed relation} & \textbf{Source TNorm} & \textbf{Target TNorm} & \textbf{Atlas CKA} & \textbf{Composed CKA}\\\midrule
Delta $\rightarrow$ Alpha & 9 & 1 & 0.620 & 0.610\\[5pt]
Delta $\rightarrow$ Beta & 9 & 10 & 0.056 & 0.038\\[5pt]
Delta $\rightarrow$ Gamma & 9 & 6 & 0.967 & 0.956\\[5pt]
\bottomrule
\end{tabular}
\end{table}
\FloatBarrier

All five prespecified response-class orderings had the predicted direction in all 11 networks. Alpha preserved more source--target alignment than Beta (mean CKA, 0.610 versus 0.038) and had higher effective rank (5.203 versus 1.224). Beta was more laminar (0.984 versus 0.585) and remained in recurrent states longer (4.402 versus 2.675 trials). Gamma exceeded the frozen CKA threshold of 0.9 in every network (mean 0.956). Each nominal exact one-sided probability was $P=0.000488$ (Figure 4b--e).

Alpha and Beta were also closer to their corresponding isolated references after composition than through the direct skip path in every network. Mean standardized feature distances were 2.156 versus 5.943 for Alpha and 5.538 versus 11.301 for Beta (three-branch Benjamini--Hochberg-adjusted $q=0.000732$ for each).

The composition transferred the predicted response classes, with state-dependent changes in their numerical properties. Of 70 control-by-descriptor checks, 57 lay strictly inside frozen reference intervals; including endpoints gave 62. Intermediate persistence transferred at one reference cycle. The two later contextual-information and delayed-geometry predictions failed their specified tests (Methods, Prospective composition). Incoming state and pathway thus influenced how the selected response classes were realized.

\begin{figure}[H]
\centering
\includegraphics[width=\linewidth,height=.55\textheight,keepaspectratio]{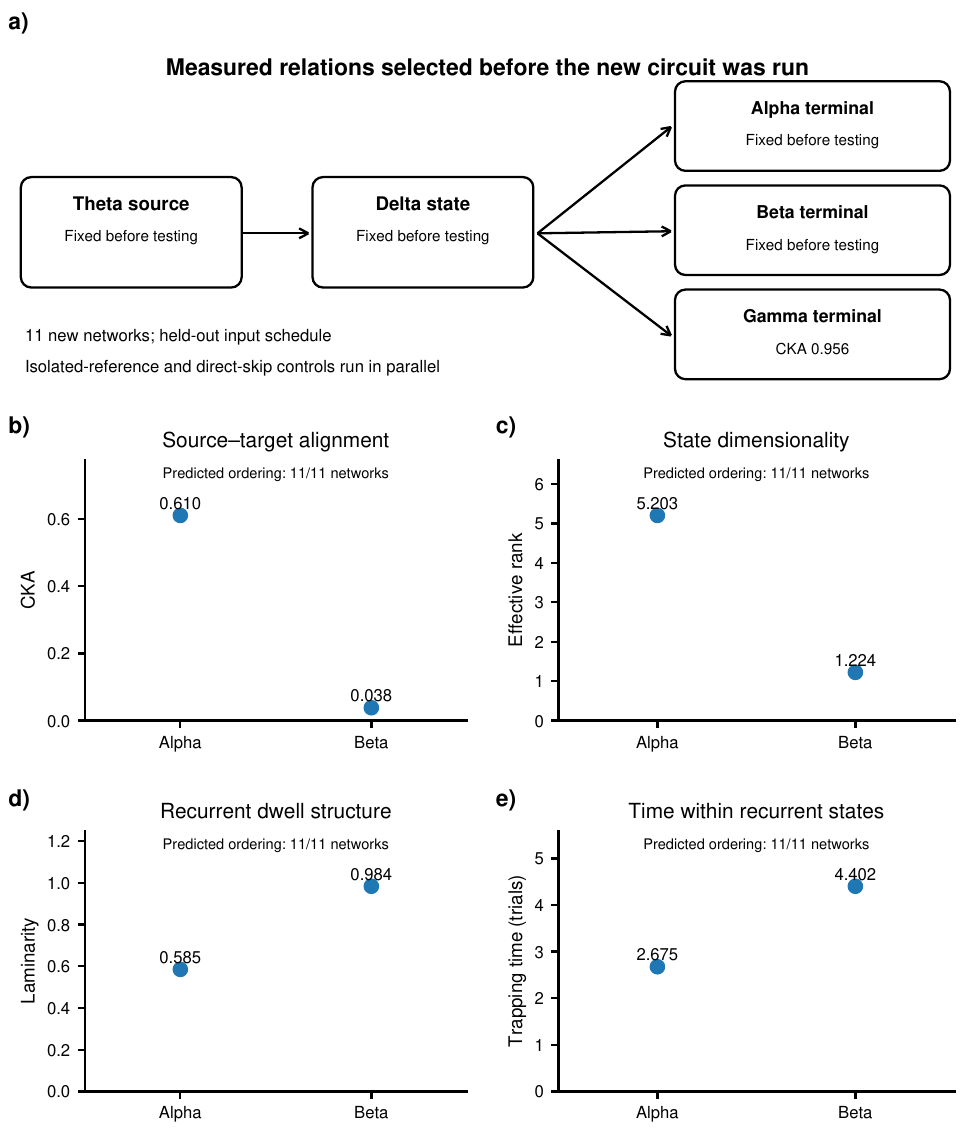}
\begin{spacing}{1.0}
\caption{\textbf{Measured interactions specify response classes in a new composition.} \textbf{a,} The frozen design uses a Theta source, a shared Delta state, and Alpha, Beta and Gamma receiving populations. Table 2 gives the selected Atlas cells and their reference values. Isolated-reference and direct-skip paths provide parallel controls. Eleven new networks received a held-out sensory schedule; operating points remained fixed throughout evaluation. \textbf{b--e,} Prospective mean source--target CKA, effective rank, laminarity and trapping time for the composed Alpha and Beta branches. Each displayed ordering had the predicted direction in all 11 networks. The fifth prediction, Gamma CKA greater than 0.9, also held in all 11 (mean 0.956). Each nominal exact one-sided $P=0.000488$. Points represent means.}
\label{fig:4}
\end{spacing}
\end{figure}
\FloatBarrier

\subsection*{Oscillations organize persistent state}
We measured state persistence, accessibility of an earlier event and contextual influence in Alpha-to-Theta, Beta-to-Beta and Theta-to-Delta. Intact and temporally averaged conditions were paired at the same network, sensory history and operating point. Schedule-preserving nulls supplied the reference for each historical measure.

In Alpha-to-Theta, mean null-centered persistence remained between 0.300 and 0.288 across one to four target cycles, compared with 0.035 to 0.012 after temporal averaging. Increasing participation under intact dynamics was associated with a more persistent, less event-specific state (Figure 5a). Oscillatory organization linked sustained activity to a stable population-state process.

Beta-to-Beta showed a reversal of the participation--persistence relationship. At one target cycle, the mean within-network correlation changed from $+0.906$ under intact dynamics to $-0.418$ after averaging, with the reversal present across all 11 networks (Figure 5b). In Theta-to-Delta, mean one-cycle persistence across the operating surface fell from 0.615 to 0.092, with reported reductions at all 121 coordinates. Its distinct persistence, event-accessibility and contextual-information landmarks also moved earlier (Figure 5c; Methods, Relation-specific persistence and historical accessibility).

Oscillatory interactions supported persistent, relation-specific states with characteristic windows of historical accessibility. Temporal averaging changed that organization while leaving some earlier labels accessible, with improved recovery at particular lags. The two conditions therefore expressed different relationships between retained information and its availability over time.

\begin{figure}[H]
\centering
\includegraphics[width=\linewidth,height=.60\textheight,keepaspectratio]{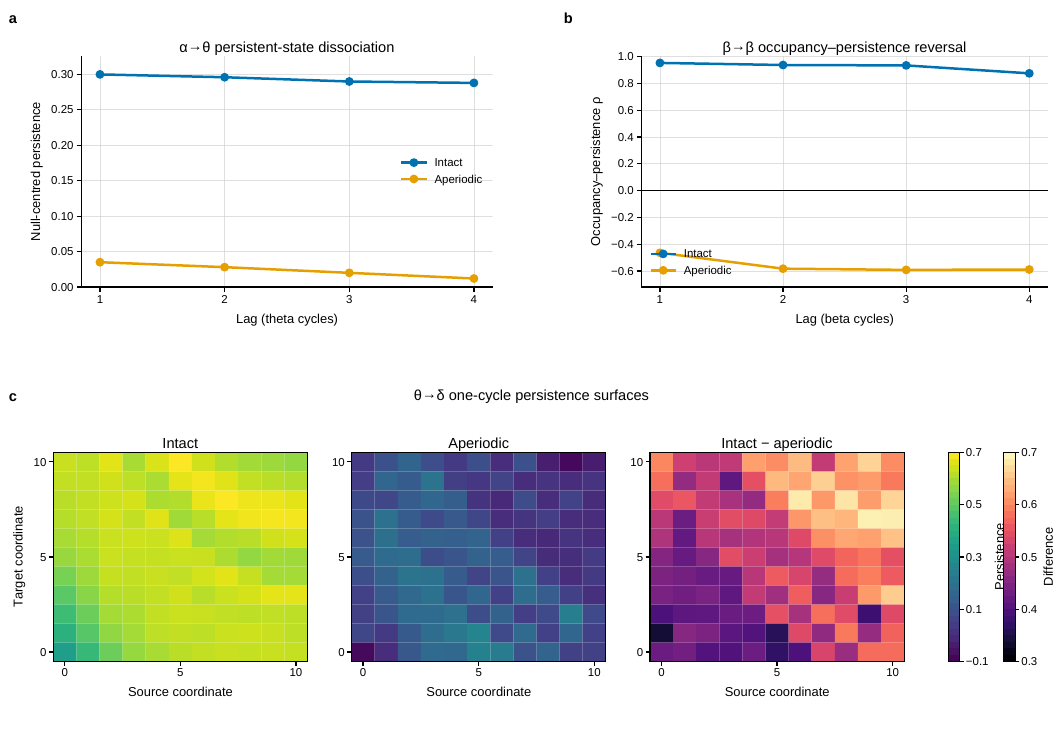}
\begin{spacing}{1.0}
\caption{\textbf{Oscillatory organization determines relation-specific persistence.} \textbf{a,} Null-centered Alpha-to-Theta persistence across one to four target cycles: 0.300, 0.296, 0.290 and 0.288 under intact dynamics, compared with 0.035, 0.028, 0.020 and 0.012 under temporal averaging. \textbf{b,} The Beta-to-Beta participation--persistence relationship is positive under intact dynamics and negative after averaging. Points summarize within-network correlations across the operating surface. \textbf{c,} Theta-to-Delta persistence surfaces at one target cycle and their difference. Each surface contains 121 measured coordinates summarized across 11 networks. One to four Theta cycles in a correspond to 50--200 model steps; one to four Beta cycles in b correspond to 15--60 steps. The Delta-cycle surface in c is evaluated at 150 steps. Figure 6 presents the separate binary-stream timing and functional-recall assays.}
\label{fig:5}
\end{spacing}
\end{figure}
\FloatBarrier

\subsection*{Historical accessibility and associative response}
A separate binary-stream assay tested the timing of historical access directly. Approximately 1,200-neuron circuits across 11 networks received the same frozen input stream under intact and averaged conditions. Causal activity summaries, chronological held-out partitions, phase-preserving label permutations and a strictly future-label control separated historical specificity from predictable features of the schedule (Methods, Historical accessibility).

A reference-period model and its harmonic improved blocked cross-validated $R^2$ over an age trend by 0.0677 under intact dynamics and $-0.0203$ after averaging. The paired difference was 0.0880, positive in all 11 networks (Holm-adjusted $P=0.00146$; Figure 6a). The history-minus-future analysis retained the joint-model condition difference (0.0886; all 11 networks; Holm-adjusted $P=0.00146$; Figure 6b). These results establish reproducible temporal structure in historical accessibility. In a secondary incremental analysis, adding the harmonic to the reference-period model improved held-out $R^2$ by 0.0252 under intact dynamics and $-0.0194$ after averaging (paired difference 0.0446; unadjusted exact one-sided $P=0.000488$).

The complementary two-event assay tested whether intact dynamics increased the schedule-adjusted difference between within-context and cross-context readout of an earlier event. The context-transfer index was 0.2031 under intact dynamics and 0.3385 after averaging, opposite to the predicted direction. The retained-past requirement passed; the complete prespecified conjunction failed. In a cue-free exploratory interval, earlier-event balanced accuracy was also higher after averaging (0.460 versus 0.293; chance reference one-third). This accuracy measures recovery of the earlier event’s explicit label by the selected offline decoder, rather than the full influence of that event on subsequent neural state. Poor label recovery can therefore coexist with detectable historical dependence, although this distinction alone does not explain the numerically below-chance value. Figure 6c presents the time-resolved historical readout from this assay. Temporal averaging improved label recovery in that interval, whereas intact organization supported the timing relationships measured in the binary-stream assay.

We examined finite associative recall in one uninterrupted, approximately 7,500-neuron circuit. Seven shape--color associations were acquired across successive experiences. All seven partial cues presented without the associated color input produced the corresponding terminal color response, including associations acquired before later experiences. The response arose at the neural outputs with fixed synaptic weights. Smaller two-object controls removed recall when teaching was withheld or memory return was cut (Methods, Finite associative recall).

The associative result establishes a functional response in one continuous network. Figure 6d summarizes the pathway schematically. Together, the memory studies show how OPU interactions organize historical accessibility and how a configured circuit uses an acquired association at its own outputs.

\begin{figure}[H]
\centering
\includegraphics[width=\linewidth,height=.63\textheight,keepaspectratio]{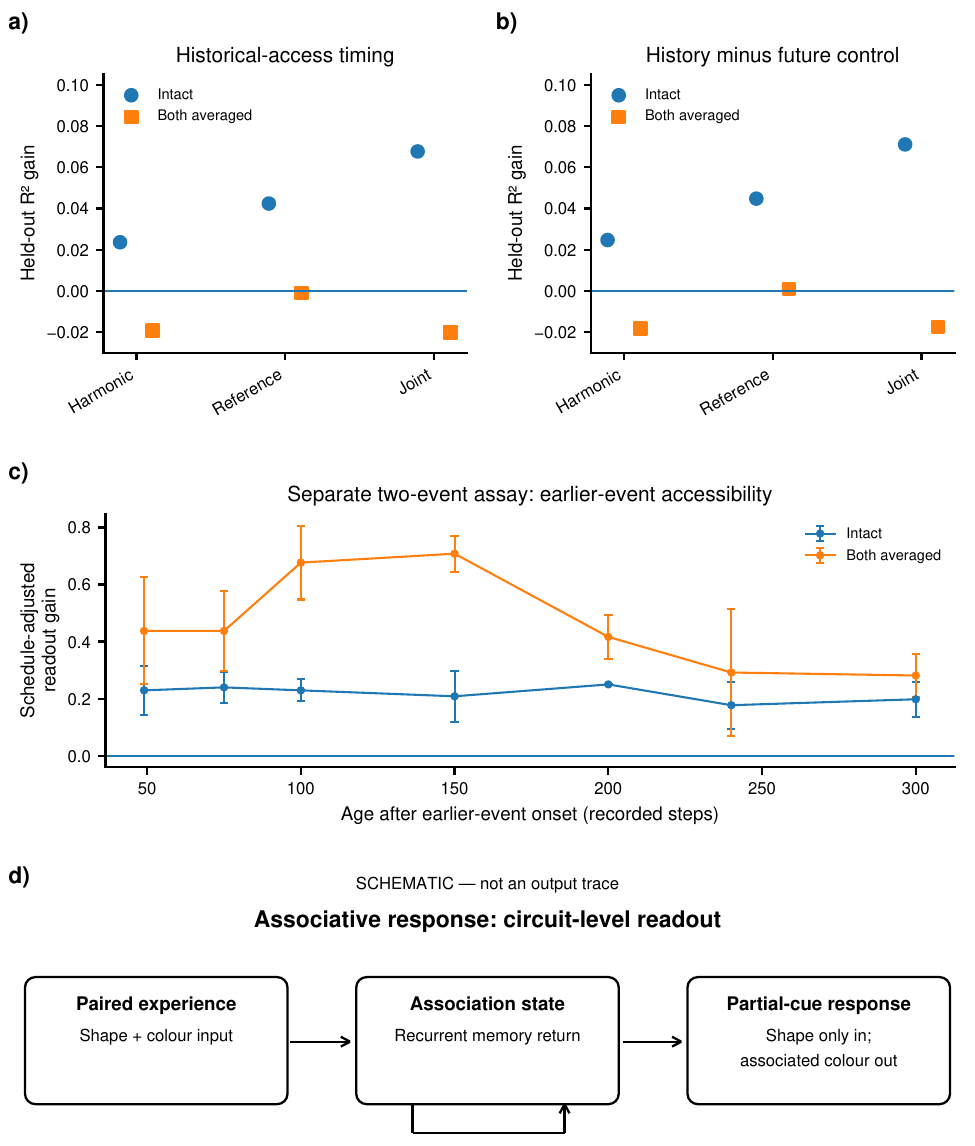}
\begin{spacing}{1.0}
\caption{\textbf{OPU circuits support historical accessibility and associative response.} \textbf{a,} Mean held-out $R^2$ improvement of harmonic, reference-period and joint timing models over an age trend in the binary-stream assay (11 paired networks). For the joint model, the intact-minus-averaged difference was 0.0880 (descriptive paired-network bootstrap 95\% interval, 0.0589--0.1168; Holm-adjusted $P=0.00146$). \textbf{b,} Corresponding history-minus-future timing gains. The joint-model condition difference was 0.0886, positive in all 11 networks (Holm-adjusted $P=0.00146$). Points in a,b are means. \textbf{c,} Earlier-event readout gain in the separate two-event assay, after subtracting a schedule-only model and evaluating off-diagonal event--context pairs. Points are means across eight paired networks; error bars are network standard deviations. Lines connect the recorded ages. \textbf{d,} Schematic of associative experience, recurrent return and partial-cue terminal response in the seven-association demonstration.}
\label{fig:6}
\end{spacing}
\end{figure}
\FloatBarrier

\subsection*{A downstream interaction converts state into action}
The controlled avoidance experiment tested whether an informative receiving state could support useful behavior. An approximately 1,400-neuron circuit received lateral obstacle information and passed it to competing motor populations controlling a virtual robot (Figure 7a). Five conditions crossed intended versus reversed sensory-to-action mapping with presence versus removal of reciprocal competition, alongside a graph-budget-matched shadow-output sham. Eight previously unseen networks were tested in three held-out worlds per condition, giving 120 runs (Methods, Controlled avoidance).

\begin{figure}[H]
\centering
\includegraphics[width=\linewidth,height=.66\textheight,keepaspectratio]{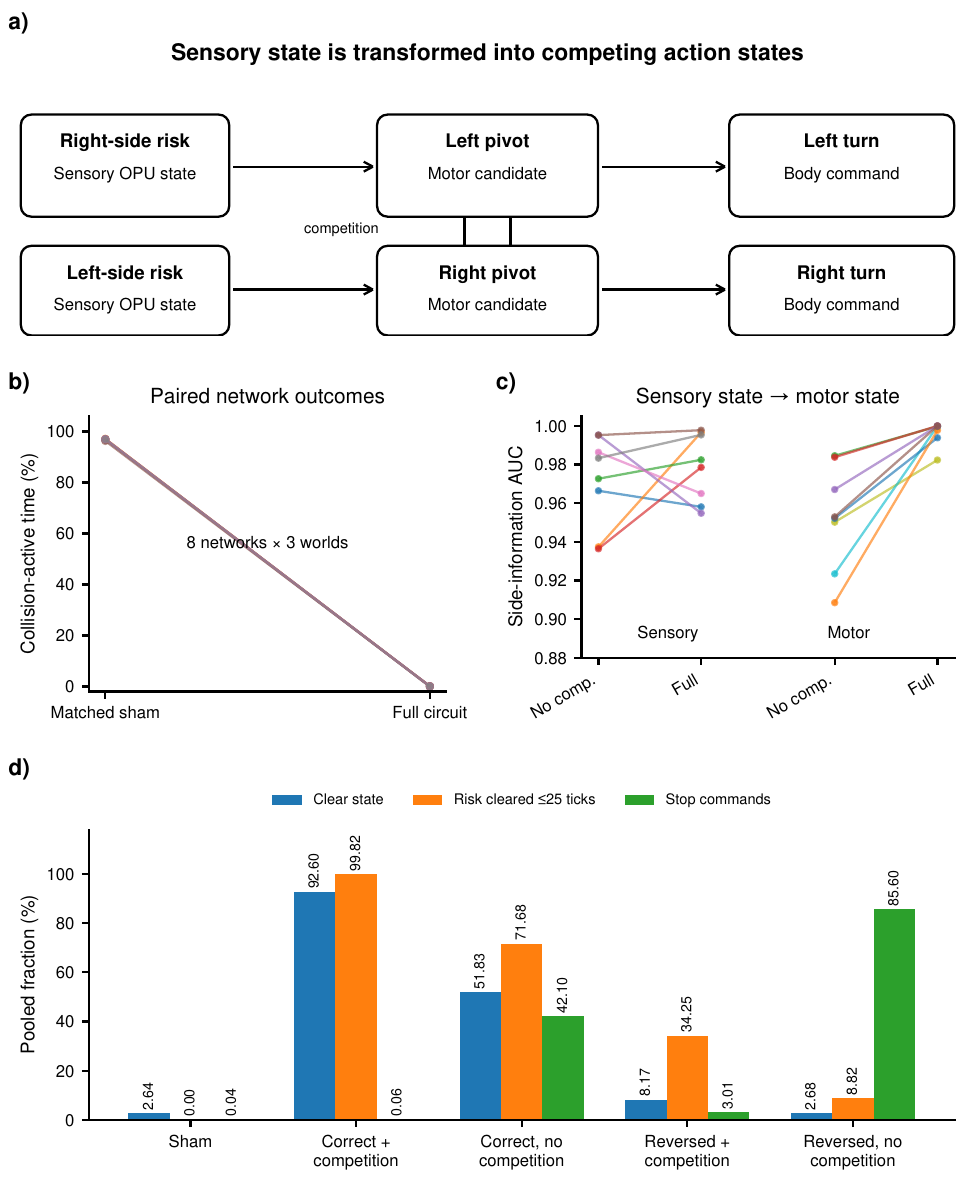}
\begin{spacing}{1.0}
\caption{\textbf{A downstream OPU interaction turns sensory state into action.} \textbf{a,} Correctly mapped circuit: right-side risk supports a left pivot, left-side risk supports a right pivot, and reciprocal suppression resolves competing motor candidates. \textbf{b,} Collision-active burden for eight paired networks, each averaged over three held-out worlds. The full circuit accumulated three collision-active ticks in 36,000 (0.0083\%), compared with 34,893 (96.925\%) in the matched shadow-output sham. The paired difference was $-96.917$ percentage points (descriptive 95\% network-bootstrap interval, $-97.044$ to $-96.733$; exact two-sided $P=0.0078125$; two-co-primary Holm-adjusted $P=0.015625$). Lines largely overlap. \textbf{c,} Paired sensory-side and motor-side AUCs under the correct mapping with and without competition. The upstream contrast gave $P=0.547$; the motor contrast was positive in all eight networks ($P=0.0078125$, secondary and unadjusted). Each condition generated its own closed-loop sensory trajectory. \textbf{d,} Pooled clear-state, risk-clearance and stop-command percentages across 24 runs per condition. Inference uses networks. Stop commands rose to 42.10\% without competition, compared with 0.058\% in the full circuit.}
\label{fig:7}
\end{spacing}
\end{figure}
\FloatBarrier

The controlled comparisons tested whether the specified motor pathway contributed to avoidance and how its directed mapping and reciprocal competition shaped behavior. In the matched shadow-output sham, selected connections were redirected to inactive destinations, disrupting the avoidance pathway’s influence on action while preserving the graph and edge budget. The full circuit accumulated 0.0083\% collision-active time, compared with 96.925\% in the sham (exact two-sided $P=0.0078125$; two-co-primary Holm-adjusted $P=0.015625$; Figure 7b). The paired difference had the same direction in every network, and the full circuit was collision-free in 23 of 24 runs.

Removing reciprocal competition increased stop commands from 0.058\% to 42.10\% and reduced pooled risk clearance within 25 world ticks from 99.82\% to 71.68\% (Figure 7d). Upstream lateral information remained similar in magnitude, whereas downstream action separability decreased (Figure 7c). These results support a contribution of competition between motor populations to resolving conflicting actions. Together, the comparisons connect the circuit’s behavior to specific directed interactions rather than merely demonstrating that neural activity reaches the motors.

A separate co-primary test evaluated a frozen scalar geometry score across 20 operating points and eight networks. The predicted ordering was reversed in every network: mean within-network rank correlation with new-circuit geometry was $-0.213$, with all eight correlations negative. The prespecified positive-direction exact test gave $P=1.000$. Across the tested configurations, higher scores were associated with lower measured geometry. The original two-co-primary correction was retained for the avoidance comparison (Methods, Companion predictive analysis; Supplementary Note 1).

Memory and avoidance establish complementary uses of the OOS. The separate memory assays measured historical accessibility and finite association. The avoidance interventions linked a downstream transformation of sensory state to the resolution of competing actions.

\subsection*{Neural sensing and decisions on the physical ANT robot}
ANT connects the architecture to real sensing and actuation. Approximately 1,400 synthetic neurons receive observations from four ultrasound sensors and a camera-based target detector. The task combines finding a requested object, avoiding obstacles and searching when the object is not visible. The external detector supplies target evidence; the neural circuit selects search, approach or avoidance, and the host application translates outputs into motion.

ANT demonstrates online sensorimotor computation through OPU interactions. Incoming observations change neural state, the dynamics resolve competing actions, and the outputs guide the next movement. The neural control circuit was configured directly, without a task-training dataset or backpropagation. Software handles sensor processing, the external camera detector, communication, motor translation and safety; action selection is performed by the neural circuit.

Supplementary Video 1 shows ANT navigating among physical obstacles. The video documents sensing and decision-making through the neural dynamics during a familiar robotics task. It is a 57-second, silent excerpt of the external-camera view from an existing demonstration recording, with frame order and playback rate preserved. The physical demonstration and controlled simulated cohort are reported separately. Additional hand and virtual-drone demonstrations are described in Methods, Embodied demonstrations.

\begin{figure}[H]
\centering
\includegraphics[width=.9\linewidth]{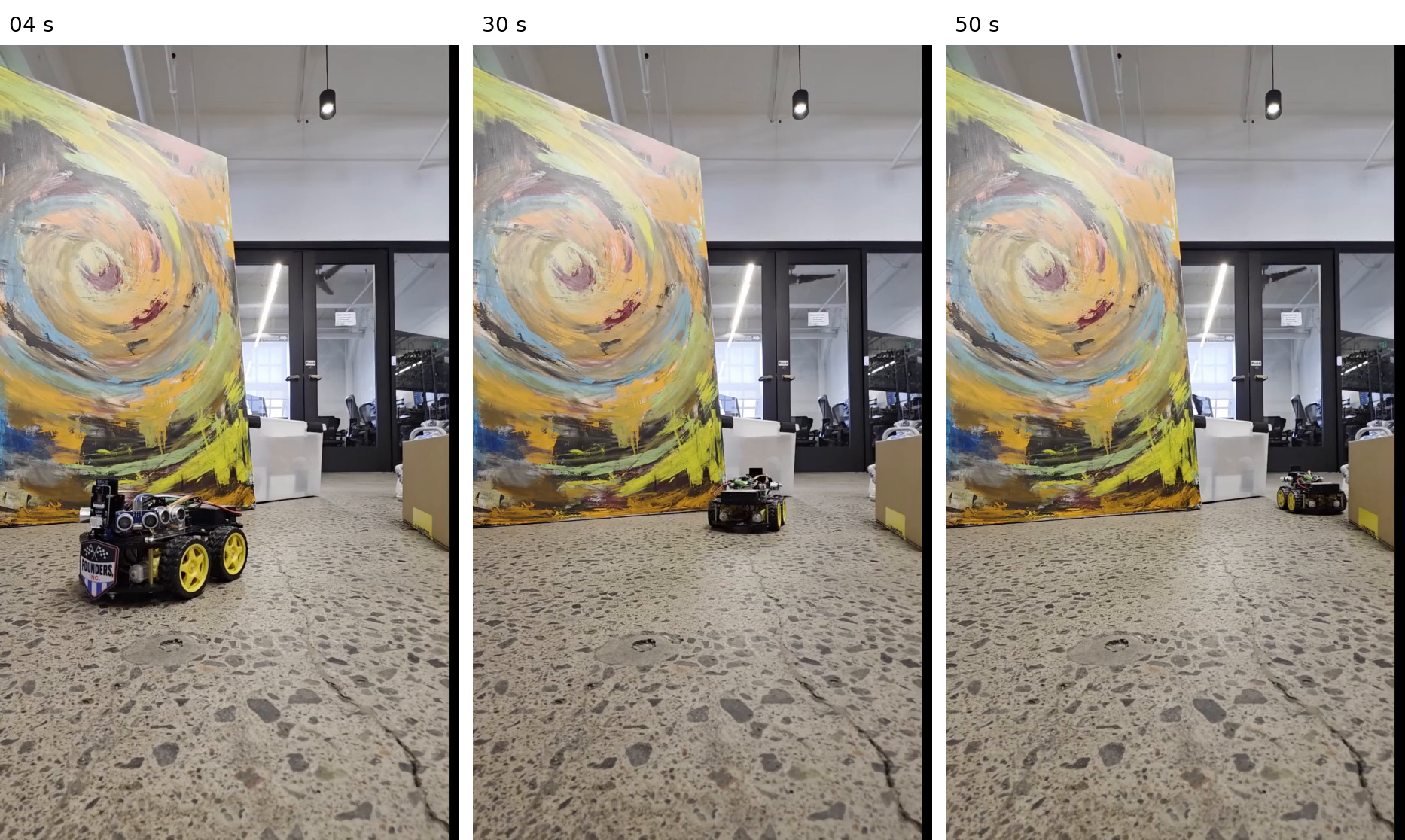}
\begin{spacing}{1.0}
\caption*{\textbf{Supplementary Video 1. Neural sensorimotor computation on ANT.} Approximately 1,400 synthetic neurons receive observations from four ultrasound sensors and a camera-based target detector and produce movement decisions. Stills at 4, 30 and 50 seconds show the physical robot. The silent, spatially cropped 57-second excerpt preserves temporal order and playback rate. The video records a qualitative physical demonstration. The MP4 is supplied as \texttt{anc/media/Supplementary\_Video\_1\_ANT.mp4}.}
\end{spacing}
\end{figure}
\FloatBarrier

\section*{Discussion}
OPUs provide the populations from which we construct artificial brains; their directed interactions provide the operations, and the OOS carries state between them. The experiments characterize this architecture from local transformations through prospective composition, memory and action. Source and receiver conditions determine families of responses that can be measured and used to select interactions for a new circuit.

Oscillatory organization determines the computational form of these interactions. Temporal averaging preserved substantial recruitment while changing state geometry, persistence and directed information relationships. Differences remained after matching mean source and target participation. The intervention therefore links periodic organization to the structure of the OOS and the timing through which historical information becomes accessible. The matching analysis controls mean recruitment; the temporal distribution of activity and effective input remain coupled to the intervention.

The prospective composition demonstrates how this characterization supports construction. Pairwise measurements specified contrasting terminal response classes before new networks were run. The shared intermediate state brought the Alpha and Beta branches closer to their intended reference classes than direct skip paths. Five directional response-class predictions transferred across all 11 new networks. The numerical and temporal checks locate the precision of that transfer: incoming state and pathway shape the particular realization of each operation.

OPUs extend a substantial tradition of dynamical computation. Liquid-state and echo-state systems use recurrent state to make present and past input available to readouts\cite{maass2002,jaeger2004}; coupled recurrent networks implement persistent states and transitions\cite{rutishauser2009}; oscillatory devices and biomimetic assemblies provide related routes to computation\cite{torrejon2017,pathak2026}. The design object developed here is the measured source--receiver interaction. Its transformation is characterized at the population interface, selected for a role in a circuit and connected to subsequent neural operations.

Fries proposes that synchronization helps determine which of several converging neuronal groups influences a receiving population.\cite{fries2015} Our framework also connects rhythm to the capacity of the receiving unit. In the OPU construction, the effective integration window constrains population size: increasing the number of incoming neurons yields progressively smaller increases in the receiving population as its rhythm-dependent capacity is approached. Rhythm therefore helps determine both how input affects the receiver and how the computational unit is constructed. Within that construction, we experimentally characterize how oscillatory organization shapes state geometry, persistence and historical accessibility. Along the same lines, Papadimitriou et al.\ describe computation through neuronal assemblies whose formation and relationships depend centrally on Hebbian synaptic plasticity.\cite{papadimitriou2020} In our reported memory and control circuits, synaptic weights remain fixed while experience changes the evolving neural state. Those state changes influence how subsequent inputs are processed. Changing synaptic strength moves an interaction through its operating range and can change the transformation it performs, as characterized through TNorm. At a fixed operating point, intrinsic temporal dynamics and activity passing between populations generate the ongoing computation.

This framework opens a route toward intelligence built through continuing interactions between experience and the present. An event changes the OOS. Later input acts on that changed state, influencing what information becomes accessible and what response follows. Memory participates in the same computation that processes new evidence. The historical-access and associative-response studies provide examples of this process with fixed synaptic weights. Experience can therefore influence subsequent computation through the evolving state of an OPU network.

In an embodied system, that process extends through the body. Neural activity produces an action, the action changes the sensory observations, and the next interaction acts on the resulting state. The controlled avoidance study connects a specific downstream relation to useful behavior. ANT makes this computation visible in a physical robot, where live sensory evidence drives neural decisions and movement. Its contribution is the realization of the sensorimotor loop through configured OPU dynamics.

The practical opportunity is to assemble increasingly capable systems from operations that preserve distinctions, combine evidence, retain context and select actions. A measured library can guide the choice of relations, while whole-circuit evaluation establishes how they work together. This makes composition a concrete research problem: identify the properties that transfer, the incoming states that preserve them and the conditions under which a circuit remains useful.

The next tests concern the range of this computational framework. Larger association sets, independently varied sensory streams, deeper compositions and external replication will establish its scalability. The XOR shortfall, context-transfer outcome and consistently inverted scalar predictor identify specific constructions and predictions requiring further work. System-level energy, latency and task-specific comparisons will establish where the architecture offers practical advantages.

The central contribution is a computational architecture expressed through OPUs and OOS: measurable interactions produce state, and that state participates in further computation. It gives us a way to construct artificial brains in which current evidence, prior experience and action remain connected throughout operation. The intelligence that can emerge from increasingly capable compositions is the next question this architecture makes possible to investigate.

\section*{Methods}

\subsection*{Observable input--output characterization}

The empirical characterization uses the stimulus sequence, configured connections and recorded output events. Each spike event is assigned to a neuron, population and model step. The binary spike pattern identifies the active assembly at that step. Participation summarizes the active-neuron fraction; assembly frequencies, dimensionality, recurrence and lag-dependent relationships describe the population-state process. Population identity, rhythm class, operating coordinate, connectivity realization and temporal condition remain attached to each observation.

Operating-response sweeps vary TNorm while retaining the specified sensory schedule and structural realization. Their ordinate is population participation, and their abscissa is the interaction's operating coordinate. The input pathway is retained with each curve. First-level responses characterize populations driven through that pathway; paired-population experiments characterize the directed transformation. Repeated cells, coordinates and lags remain nested within the independently generated network.

The interventions manipulate phase-dependent sensitivity and periodic release. Phase averaging holds the sensitivity modulation at its cycle mean; release averaging holds the periodic release modulation at its cycle mean. Their combination applies both substitutions. Recruitment, state geometry, recurrence, source--target relationships and persistence are the measured outcomes. Paired comparisons retain the same network and input schedule; the secondary relation analysis matches mean source and target participation.

\subsection*{Population-state measurements}

For a population of $n$ neurons observed over $T$ model steps, let $X\in\{0,1\}^{T\times n}$, where $X_{ti}=1$ when neuron $i$ emits an event at step $t$. The row $x_t$ is the instantaneous active assembly. Its active-neuron fraction and the raw mean occupancy are
\begin{equation}
\rho(t)=\frac{1}{n}\sum_{i=1}^{n}X_{ti},\qquad
\bar\rho=\frac{1}{Tn}\sum_{t=1}^{T}\sum_{i=1}^{n}X_{ti}.
\end{equation}
Writing $p_{\mathrm{act}}$ for the fraction of steps containing at least one event and $a$ for mean occupancy conditional on activity gives
\begin{equation}
p_{\mathrm{act}}=\frac{1}{T}\sum_{t=1}^{T}\mathbf{1}[\rho(t)>0],\qquad
a=\frac{\sum_t\rho(t)}{\sum_t\mathbf{1}[\rho(t)>0]},\qquad
\bar\rho=p_{\mathrm{act}}a.
\end{equation}
The conditional mean is evaluated when active observations are present. The release gate is the periodic modulation determining when retained input contributes to a neuron's response. Gate participation expresses cycle-averaged occupancy relative to the configured release fraction. Let $\eta_B$ denote that fraction for population $B$ and let $\langle\rho_B\rangle_{C_B}(t)$ denote occupancy averaged over one nominal cycle. The unclipped rhythm-normalized measure is
\begin{equation}
g_B^{\circ}(t)=\frac{\langle\rho_B\rangle_{C_B}(t)}{\eta_B}.
\end{equation}
The normalization remains fixed across the paired temporal conditions. Raw occupancy measures activity per recorded step; the normalized measure expresses that activity relative to the release fraction. Nested, activity-conditioned and ablation-matching analyses use the unclipped measure. Figure 3a displays the clipped form of this measure. These curves are interpreted alongside the activity-matched comparisons and state measurements.

For the observation set $\mathcal T$ used by an exact-state analysis, the empirical probability of assembly $s$ and its Shannon entropy are
\begin{equation}
\widehat p(s)=\frac{1}{|\mathcal T|}\sum_{t\in\mathcal T}\mathbf{1}[x_t=s],\qquad
H=-\sum_{s:\widehat p(s)>0}\widehat p(s)\log_2\widehat p(s).
\end{equation}
The exact-state entropy values in Figure 3b use active steps, $\mathcal T=\{t:\rho(t)>0\}$. Other endpoints retain the observation set recorded by their analysis.

Trial states are per-neuron activity means over each input presentation. Historical state measurements use the causal trailing summaries defined for the corresponding assay. For a state representation $Z$ with nonzero singular-value power, let $\sigma_j$ be its singular values. The variance-based effective rank is
\begin{equation}
q_j=\frac{\sigma_j^2}{\sum_k\sigma_k^2},\qquad
R_{\mathrm{eff}}=\exp\!\left[-\sum_{j:q_j>0}q_j\ln q_j\right].
\end{equation}
Thus each analysis retains its specified state representation before evaluating the singular-value power spectrum.

For aligned source and target representations $U$ and $V$ with $m$ corresponding observations, define $J_m=I_m-\mathbf{1}\mathbf{1}^{\mathsf T}/m$, $U_c=J_mU$ and $V_c=J_mV$. Linear centered kernel alignment is\cite{kornblith2019}
\begin{equation}
\operatorname{CKA}(U,V)=
\frac{\|U_c^{\mathsf T}V_c\|_F^2}
{\|U_c^{\mathsf T}U_c\|_F\,\|V_c^{\mathsf T}V_c\|_F},
\end{equation}
for nonzero denominator. Source and target populations can have different numbers of neurons; their rows refer to aligned observations. Native-lag persistence applies this measure to a receiving representation and its lagged counterpart, then subtracts the mean under the schedule-preserving null described below.

\subsection*{Execution platform and experiment interface}

The studies use OPU circuits executed on earlier reference runtimes and the Prima-Lux platform. Each result remains associated with its study configuration and execution settings. The experimental interface supplies the input sequence, initial-state conditions and temporal variant and records the neural outputs. Synaptic weights were fixed in the reported memory and control assays; neural and short-term synaptic states continued to evolve.

A network denotes one independently generated connectivity realization of a specified circuit design. Paired conditions use the same realization. Supplementary Table 1 summarizes the study designs and experimental units. Operating cells, neurons, lags, episodes and worlds are repeated observations within the relevant network. Each historical-access and two-event assay used a shared sensory sequence across networks.

Model steps index the ordered neural computation. Sensory episodes and world ticks identify their respective stimulus and environment schedules. The cycle periods reported here specify algorithmic time. Hardware execution cadence depends on the runtime and workload.

The physical hand and ANT robot use real sensing and actuation; Webots supplies the virtual drone body and flight environment; the controlled avoidance study uses a virtual robot. The host application supplies observations and translates neural outputs into actuator commands. In object seeking, the camera detector supplies target evidence and the OPU circuit selects the action.

Prima-Lux provides up to 10,000 synthetic neurons on FPGA-based hardware. A Python SDK, visual designer, output streaming and neural telemetry expose the experimental interface (\url{https://github.com/artificial-brains-inc/artificial_brains}). Neural computation runs locally; preparing a new design uses the online build service. The historical results retain their recorded study runtimes and execution settings.

\subsection*{Factorial characterization and Atlas}

Four source-only families and all 16 ordered pairings of four rhythm classes were tested. Eleven-by-eleven operating grids yielded 1,936 paired cells and 21,296 network-level conditions across 11 networks. The broader pipeline comprised 220 graph runs and 28,336 population instances.

Measurements included firing participation, exact-state entropy, effective rank, recurrence and source--target similarity. Centered kernel alignment\cite{kornblith2019} and normalized mutual information described relationships between recorded states. The occupancy-only analysis model included source and target firing participation and their quadratic and interaction terms. The full model also included source and target operating coordinates and their quadratic and interaction terms.

Two validation questions are distinguished. The mathematical Atlas evaluated pair-specific response surfaces by holding out complete source rows and complete target columns, reporting the lower of those predictive scores as the conservative result. Its 240 pair-by-descriptor comparisons comprised 16 pairings and 15 descriptors. Occupancy-dominant relationships required conservative $R^{2}$\ensuremath{\geq}0.70 and less than 0.10 improvement from coordinates; coordinate-conditioned relationships required full-model $R^{2}$\ensuremath{\geq}0.70 and improvement of at least 0.10. The separate source-contribution analysis compared target-only, source-plus-target and interaction models with complete networks held out. The validation schemes test generalization across operating coordinates and across network realizations, respectively. Within-run significance was assessed using Monte Carlo randomization, with 10,000 schedule-preserving permutations where that null was available. Complete counterbalanced design rows were reassigned across neural epochs while preserving each row’s order, positions and stimulus blocks. The population and transfer analyses used 5,000 bootstrap resamples, retaining temporal or epoch blocks rather than resampling individual neural events. Maximum-statistic correction across each complete 11 \ensuremath{\times} 11 surface controlled the corresponding within-run multiple-comparison family. These procedures tested whether observed geometry or information measures exceeded structure expected from the retained schedule and sampling design.

Rhythm-relative history analyses also used a finite null consisting of all 17 non-zero whole-epoch shifts of the 18-epoch schedule. This was an enumerated null, distinct from the 10,000-draw Monte Carlo analyses. Across networks, effects were aggregated within each network and evaluated using all 2,048 sign assignments for the eleven paired networks. Benjamini-Hochberg adjustment\cite{bh1995} was applied within the declared analysis families. Independent connectivity realizations supplied replication; operating cells, neurons and lags supplied repeated measurements. The experimental protocols were fixed before the corresponding confirmation analyses.

The fitted pair-specific surfaces predicted target firing participation with conservative held-out $R^{2}$ values from 0.9792 to 0.9995. Across the 240 pair-by-descriptor relationships, conservative cross-validated $R^{2}$ averaged 0.487 for occupancy-only models and 0.724 when source and target operating coordinates were retained. Of those relationships, 54 were occupancy-dominant, 104 were coordinate-conditioned, 48 were moderately represented by the fitted Atlas and 34 required empirical lookup. These are the numerical comparisons summarized in Results and classified in Supplementary Figure 1.

The neural-dynamics and behavioral Atlas organizes measured state descriptors by directed rhythm pair, source and receiver operating points, and input pathway. Descriptive labels summarize passing, persistence, reorganization, recruitment and temporal relationships. Table 1 links the measured properties to their experimental evaluations; logical, fuzzy, memory and motor operations are assessed in their respective circuits.

\subsection*{Five-rhythm operating maps and input-path comparison}

The fifth-rhythm extension added the Gamma-only family and the nine missing directed relationships involving Gamma. Together, the source-only and paired families comprised 30 graph designs under 11 independently generated networks: 330 runs, 25 paired surfaces and 3,025 paired operating cells. Source and target TNorm ranged from 0 through 10 at integer coordinates. The sensory schedule contained 18 counterbalanced epochs of nine joint conditions, yielding 162 trials. Complete networks, rather than individual operating cells, supplied independent structural replication.

Figure 2 uses the Gamma-mediated input-path measurements. Each operating-cell entry is the across-network summary for that coordinate. The two pair-level maps show medians over the 121 coordinate summaries for each directed pair. Independent structural replication is supplied by the 11 networks. Source Data supply the pair-level medians and the three operating maps shown in Figure 2.

CKA compares centered representations through normalized kernel alignment; NMI summarizes the normalized mutual information between the recorded source and target representations. Effective rank describes the dimensionality of the population-state representation. Determinism measures diagonal organization in recurrence analysis; laminarity and trapping time describe repeated residence in similar states, with trapping time expressed in sensory-trial units. These descriptors summarize the recorded population states. Table 1 identifies the measurements and their supporting experiments. Complete circuits are evaluated for their intended operations.

The input-path comparison used the same 30 graph families and 11 network realizations with the sensory relay projecting directly to the first rhythmic population rather than through the Gamma stage. The comparison changes the incoming state and architecture-specific allocation. Recruitment curves were considered identifiable only with fit $R^2\geq0.95$ and empirical recruitment range at least 0.05; 108 of 150 curves met that criterion. Realized-state matching used source and target participation, not CKA, to select corresponding coordinates between input paths. CKA remained the evaluation variable. Across 3,025 matched cells, the reported prediction changed from $R^2=0.909$, mean absolute error 0.0530, to $R^2=0.939$, error 0.0357. For the 1,089 Delta-involving cells, the corresponding values were 0.934 and 0.0515 versus 0.964 and 0.0321. Flat or saturated recruitment curves were not extrapolated. Supplementary Figure 2 reports the aggregate input-path comparisons.

\subsection*{Oscillatory organization and activity-matched comparisons}

This paired ablation is distinct from the frozen-profile analysis below. Four first-level rhythm families and seven directed two-level relations were assessed across 11 structural networks. The first-level factorial used intact timing and two separate temporal alterations plus their combination: 44 intact runs and 132 altered runs, or 176 graph--condition--network observations. Seven directed relations were then compared under intact and both-averaged conditions: four anchor relations (Delta-to-Alpha, Delta-to-Delta, Delta-to-Theta and Theta-to-Theta) and three history-focused relations (Alpha-to-Theta, Beta-to-Beta and Theta-to-Delta). Their 77 altered runs reused 77 intact references. Across these stages, there were 209 new altered runs and 121 reused intact runs. Shared controls retain their original network identities in all paired analyses.

Phase averaging replaces the within-cycle modulation of neuronal sensitivity with its cycle mean while retaining periodic release. Release averaging replaces the periodic modulation of when retained input contributes to the neuronal response with its cycle mean while retaining phase-dependent sensitivity. The combined condition applies both substitutions. Population identity, structural connections, synaptic weights, remaining state dynamics, sensory events and presentation order were retained. Gamma source populations retained their original temporal dynamics. The condition key is I, intact; P, phase mean; R, release mean; A, both averaged. The combined condition is also termed aperiodic in the manipulated populations.

Recruitment, exact-state entropy, effective rank, recurrence, source--target alignment and information were analyzed separately. CKA quantifies alignment of the measured representations; normalized mutual information (NMI) summarizes information shared by their assigned states. Fusion and fragmentation describe convergence or splitting of those assignments. Each descriptor retains its own measurement definition. Figure 3 reproduces the measured plots and reported means.

Condition comparisons were paired at identical network, graph, population or directed relation, operating point and lag. First-level factorial main effects and interactions were summarized within network. Across-network inference used all $2^{11}=2{,}048$ sign assignments with Benjamini--Hochberg adjustment within the original metric families. Within-run analyses used schedule-preserving Monte Carlo nulls and block bootstraps as specified under Statistics and reproducibility.

A secondary activity-matching analysis rematched intact and both-averaged cells within each anchor relation and network, allowing operating coordinates to differ. Both source and target participation had to differ by no more than 0.03. Median matched-pair differences were evaluated across the 11 networks. This analysis controls mean recruitment. Spike-time distributions and realized effective input remain free to vary between conditions.

\subsection*{Relation-specific persistence and historical accessibility}

The relation-specific history analysis used Alpha-to-Theta, Beta-to-Beta and Theta-to-Delta from the paired ablation. The frozen sensory schedule comprised 18 complete nine-condition epochs and 162 trials. Every joint condition appeared twice at every sequence position, ordered distinct-condition pairs were counterbalanced within epochs, and immediate repetitions were excluded. Intact and altered observations were paired at identical network, graph, operating point and lag.

The recorded state was summarized with causal trailing activity features. Target-state persistence measured similarity to an earlier state of the same population; past-event geometry related current state to a preceding event label; contextual information measured association with a preceding event conditional on the current event. Each measure was corrected against its specified schedule-preserving null. Native lags are expressed in target-cycle units.

For Alpha-to-Theta and Beta-to-Beta, the complete operating surface supplied a participation--persistence correlation within each network at one to four target cycles. Participation--event and participation--decoding relationships were separate companion measurements. Theta-to-Delta was examined across the complete surface, at prespecified history landmarks, and within a fixed high-drive region for the ordering of persistence, event-accessibility and contextual-information peaks. Forty-two boundary-selected cells shared by both conditions supplied a targeted extended-lag comparison. Cells and lags are repeated observations within networks. Peak summaries accompany the coordinate-level paired effects.

Within-run temporal nulls enumerated the 17 nonzero whole-epoch shifts of the 18-epoch schedule, preserving its order and within-trial phase while shifting neural-history alignment. Other current-content and trial-relative analyses used 10,000 schedule-preserving Monte Carlo permutations and 5,000 block-bootstrap draws. Across-network contrasts used exact paired sign flips over 2,048 assignments and Benjamini--Hochberg correction within the original temporal families. Early-lag increases in direct past-label accessibility remain part of the result. The inferential endpoints describe the form and timing of persistent state.

Figure 5 reports persistence with native-cycle-relative lags. The source tables provide the network-level Alpha-to-Theta persistence curves, Beta-to-Beta participation--persistence correlations and the 121-cell Theta-to-Delta surface. The nominal periods specified in Computational model convert these lags to model steps within their respective assays.

\subsection*{Temporal perturbations}

Three prospectively specified relations were tested under intact timing, phase averaging, release averaging and their combination, using the intervention definitions above. Graph identity, synaptic weights and stimulus delivery were retained. Unaffected components retained temporal structure; the combined condition is called temporally averaged or aperiodic. Conditions were applied through runtime variants. Realized firing and effective input were measured in each condition. Ninety-nine new altered runs reused 33 intact controls, yielding 132 relation-condition-network combinations and 15,972 cell rows. Controls retain their original network identities in the paired analysis.

The primary endpoint was loss of posterior support for a frozen state profile within previously assigned cells. Classifier inputs were entropy, effective rank, determinism, laminarity, trapping time, participation and occupancy; source-target similarity was excluded. The classifier remained frozen during evaluation. Full-surface dimensionality, activity and dwell summaries were secondary.

Each of nine relation-by-alteration contrasts used the mean within-network posterior loss and an exact one-sided sign-flip test over 2,048 assignments. The replication rule required at least nine positive networks and Benjamini-Hochberg-adjusted q\ensuremath{\leq}0.05 across nine tests.\cite{bh1995} All nine comparisons had eleven positive effects. The profile assignment, scaling and classifier parameters remained frozen throughout the perturbation analysis. The secondary factorial contrasts separated the contributions of the two alterations and their interaction. Supplementary Figure 3e plots the mean posterior losses in Supplementary Table 2 without refitting the classification. Supplementary Figure 3b--d uses the recorded network means over all 121 operating cells for the Theta-to-Theta relation. These full-surface descriptive measurements are distinct from the posterior-loss endpoint restricted to the originally assigned profile cells. The source-data table contains all 132 relation-condition-network rows, allowing the corresponding descriptors to be examined for each of the three relations.

\subsection*{Historical accessibility}

The binary-stream historical-access assay presented 1,000 independently generated binary color symbols, each for 25 recorded steps, giving 25,000 steps per run. The same frozen stream was used in both conditions and all 11 networks. Three preselected receiving configurations were observed; their endpoint values were aggregated by the within-network median, not treated as independent networks. Neural features were causal trailing activity summaries, excluding future samples.

The phase-controlled analysis used 819 anchor symbols. Chronological partitions assigned 371 anchors to training, 123 to validation and 125 to testing, with two 100-symbol embargo intervals separating these partitions. Training-derived feature standardization was applied to subsequent partitions. Ridge regularization was selected from powers of ten between $10^{-6}$ and $10^{2}$ using validation area under the receiver-operating-characteristic curve (AUC); ties favored the smaller penalty. The chosen model was refitted using training plus validation observations and evaluated on untouched test anchors.

Historical excess was past-label AUC minus mean AUC under a Monte Carlo null of 200 held-out label permutations within six reference-phase classes. Models, scores, partitions and phase composition remained fixed. The future control used the first symbol beginning strictly after readout; onset exactly at readout counted as current. This comparison evaluates historical specificity.

Over ages 50--2,500, a quadratic age trend was compared with trend plus sine/cosine terms at the reference period, its half-period harmonic, or both. Eight contiguous age blocks defined cross-validation; $R^{2}$ used concatenated held-out predictions. Blocks contain repeated ages. The history-minus-future contrast subtracts the corresponding predictive improvements.

The three timing models were selected from intact data and frozen before the averaged-condition comparison. The altered-condition comparison was prospective; the intact data supplied the discovery analysis. Paired one-sided sign-flip tests and Holm correction\cite{holm1979} were retained within each declared three-model family. The reported 0.0880 condition difference had a descriptive paired-network bootstrap interval of 0.0589--0.1168. The incremental harmonic comparison retains its secondary status and nominal probability.

Excluding a configuration with degenerate phase-centered variance retained the principal temporal directions. A newly discovered averaged-condition timescale remained exploratory and potentially window- or cadence-dependent. Phase and event age remained coupled in this design. The analysis measures timing structure in the historical-readout signal.

\subsection*{Context-dependent historical readout}

The two-event contextual-memory assay used eight networks in two paired conditions, for 16 continuous runs. Each contained nine epochs of nine episodes: 81 episodes, 36,450 steps per run and 583,200 steps in total. The event library comprised red circle, green triangle and blue square. The same three events served as earlier event and later context, yielding nine ordered pairs. Each event paired one shape with one color.

Episodes presented the earlier event at steps 0--19 and context at 50--69, with blank input otherwise through step 449. Every pair occurred once per epoch and at every sequence position across epochs. State continued between episodes, including during blank-input intervals. All networks shared the sequence and intact-then-averaged condition order.

Epochs 0--4 supplied training observations, 5--6 validation and 7--8 testing. Global readouts therefore had 45, 18 and 18 episode observations at a given within-episode time; off-diagonal primary evaluation had 12 test observations. All nine pair classes remained available during training. Features used fixed causal trailing activity windows; the receiver analysis used an 80-step window. Ridge penalties were 0.01, 0.1, 1, 10 and 100, selected by validation balanced accuracy, with ties favoring the larger penalty. This tie rule differs from the one used in the binary-stream assay. Fitting and standardization used no test observations.

Balanced accuracy is mean class recall. Three-event and nine-pair global readouts have class-chance references of one-third and one-ninth, respectively. Classifiers were fitted separately by network, condition, age and target. Each neural-plus-schedule model was compared with a schedule-only model on the same observations. Available covariates included local sequence position, the preceding pair and the non-target event only after presentation; absolute epoch and episode identities were excluded. A target being predicted was not directly supplied as its own nuisance feature.

The context-transfer index contrasted schedule-adjusted within-context and cross-context readability. A large value can reflect poor transfer as well as informative within-context state. A context-centered control subtracted means estimated from training observations before repeating the analysis. The prespecified primary contrast was intact minus averaged at lag 240 from earlier-event onset, not a post hoc search over favorable ages.

The prespecified context-transfer claim required five conditions jointly: a positive primary contrast, positive retained-past gain in intact, a positive context-centered contrast, acceptable future-context association before its presentation, and acceptable inter-episode carry-over. Positive gates required a positive mean, at least seven of eight positive network values and one-sided exact P\ensuremath{\leq}0.05. Each validity control required mean gain no greater than 0.05 and a non-significant positive-direction test in both conditions. The retained-past requirement passed; the complete conjunction did not. The primary index means were 0.2031 and 0.3385, giving an intact-minus-averaged difference of $-0.1354$, with three of eight network differences positive (one-sided exact $P=0.918$). Context centering yielded an intact-minus-averaged difference of $-0.1927$, with two of eight positive differences. This contrast evaluates the prespecified context-transfer prediction.

The exploratory cue-free interval comprised ages 80--380 after the final context sample. Every sample contributing to the receiver feature window then occurred under blank external input, before the next episode. Metrics were averaged over the 301 ages within each network before comparing the eight paired networks. The feature windows exclude cue-period activity; ongoing state retains preceding-episode influences. The interval analysis used 100,000 paired-network percentile-bootstrap draws and exact two-sided sign flips; Holm correction covered eleven non-redundant descriptors within each interval. Pointwise estimates describe the recorded trajectory; the corrected interval analysis supplies inference.

Transition-lag analyses used 2,000 Monte Carlo history-label permutations within epoch-by-context strata, retaining the sensory sequence, split membership and context labels. The full and schedule-only models were refitted for each permutation, including validation-based regularization selection. The pointwise probability was the number of null statistics at least as large as the observed statistic, plus one, divided by 2,001. This plus-one correction avoids a zero probability from a finite random sample. Exact network-level comparisons were Holm-adjusted across the 17 examined lags within each temporal condition.

The dense-age analysis used a different null: 2,000 permutations of held-out labels within test epoch, with fitted models and predictions held fixed. Its plus-one probabilities evaluate the fixed held-out predictions, rather than the refitted conditional-label test. These distinct nulls were kept separate. The interval analysis used 100,000 paired-network bootstrap draws and preserved network pairing throughout. Recorded input audits, telemetry hashes and reconstructed neural-event checks tied the analysis to the completed runs. The two-event contextual-memory assay complements the binary-stream timing assay by measuring a different aspect of history-conditioned state. Figure 6c plots the off-diagonal earlier-event readout gain: balanced accuracy of the neural-plus-schedule model minus that of the schedule-only model, summarized across the eight networks at seven recorded ages. Error bars are network standard deviations. Lines connect the measured ages. The trajectory describes the two-event historical readout.

\subsection*{Finite associative recall}

The finite associative-recall test used one network and seven shape–color associations in an uninterrupted session. Associations were presented through paired shape and color inputs; recall was assessed with shape input while external color input was absent. Synaptic weights remained fixed. A response counted as correct only when the expected terminal color output was active and competing color outputs were inactive. The answer was scored directly from neural terminal activity.

Smaller two-object controls reversed the experienced colors, withheld teaching or cut memory return; they were assessed separately from the seven-association session. Blank activity was not scored as recall. The reported result is finite associative recall within one continuous network.

\subsection*{Operations and fuzzy logic}

The structured-input operations study comprised 152 technically valid runs across eight networks and two tested substrates. Independently supplied shape and color carriers provided operand streams. Addition was assessed against realized neural operand magnitudes, including leave-one-identity-out prediction. The analysis fitted output against those measured operands; leaving out a sensory identity tested whether the fitted relationship transferred to that carrier. The reported means across eight networks were 0.954 fitted and 0.953 held out for one construction, and 0.903 fitted and 0.892 held out for the other. Signed comparison was assessed by a signed-response fit and sign agreement. OR, AND, veto and exclusive-OR were evaluated over the tested input combinations using the study’s strict output-separation criterion.

The fits and truth-table counts summarize the completed engineering panel. Identity cases within a network are dependent. OR, AND and VETO achieved strict separation in every tested identity/network combination for one substrate. The reported signed subtraction fit was $R^{2}$ = 0.940 with 93.4\% sign agreement; the tested XOR construction met the strict criterion in 45 of 72 identity cases. The analysis used fixed thresholds, scoring windows and technical-validity rules; the associated configuration and per-trial records are retained with the study. Error analysis localized the main XOR shortfall to the both-active condition and identified unequal effective source amplitudes as a candidate explanation. A correction based on matching source reference levels was proposed, not demonstrated in this panel.

A separate distributed-fuzzy engineering validation tested five graded input levels from absent to full support across three networks. Fuzzy OR, NOT and AND reproduced the expected output patterns at those tested coarse levels, as did one downstream reuse of a complemented input with an OR operation. The validation assessed complete neural output patterns, not a scalar response supplied by software. Finer levels and more deeply composed cases remain separate validation tasks. This engineering record is not pooled with the 152-run arithmetic/Boolean study and supplies no additional formal significance test.

\subsection*{Decision-layer studies}

The temporal-composition panel varied two connection motifs and two six-level timing factors, giving 72 active configurations. Each discovery configuration received two sessions for each of two histories, totalling 288 fresh sessions. Ten selected candidates were subsequently tested with four sessions per history and a matched zero-history-coupling sham. The selected configuration’s eight-of-eight mapping and sham’s five-of-eight mapping are session counts, not eight independent networks. Selection and the small confirmation sample limit generalization.

The separate shallow-action assay comprised one completed trial measuring raw outputs, rule agreement, support ordering and preservation of ties. Normalization was used for output diagnostics. Unlike the avoidance circuit, this panel lacked winner-take-all competition. Rule decisions, support ordering and ties were scored at the neural output rather than converted into a software-selected action.

\subsection*{Prospective composition}

Reference intervals, operating coordinates and response-class predictions were frozen from the five-rhythm pairwise Atlas before evaluating a new three-stage graph. Eleven previously unused connectivity realizations, R11--R21, received a held-out counterbalanced sensory schedule. Each run supplied 162 trials of 20 recorded steps, giving 3,240 analyzed steps per network. Labels were assigned from recorded time indices; an extra terminal callback was excluded. No coordinate sweep, neural-policy training or post-observation retuning was performed.

The prospectively selected pathway used Theta at TNorm 10, a shared Delta intermediate at TNorm 9, and Alpha, Beta and Gamma terminal populations at TNorm 1, 10 and 6, respectively (Table 2). Operating points were frozen before testing. The base interaction, three composed terminal paths, three isolated Delta-to-terminal references and three direct Theta-to-terminal skip paths supplied ten analyzed branches. Each receiving population was constructed for its immediate source. Seven control branches supplied the base and isolated/skip comparisons. Branch measurements were repeated observations within each network.

The five directional endpoints were Alpha CKA greater than Beta CKA, Alpha effective rank greater than Beta effective rank, Beta laminarity greater than Alpha laminarity, Beta trapping time greater than Alpha trapping time, and Gamma CKA greater than 0.9. The reported exact one-sided probabilities enumerate all 2,048 network-level sign assignments. Figure 4 shows the means and 11-of-11 directional consistency. The five reported probabilities are nominal.

Calibration compared the means of ten descriptors in each of seven control branches with the frozen 2.5th and 97.5th percentiles. Strict interior inclusion gave 57 of 70 checks; inclusive endpoints gave 62. Zero-width reference intervals account for the five-check difference. These 70 checks describe calibration, not 70 independent replications. Standardized distances used seven fixed descriptors relative to the corresponding isolated reference. Alpha and Beta each improved in all 11 networks (exact one-sided $P=0.000488$; three-branch Benjamini--Hochberg-adjusted $q=0.000732$). Gamma improved in nine networks ($P=q=0.00342$), but near-zero reference variance made its absolute distance magnitude unstable, so that magnitude is not interpreted or plotted.

Three separately specified temporal-reference checks evaluated intermediate-state persistence, contextual information and delayed source--target geometry at one, two and four reference-cycle multiples, respectively. The persistence mean, 0.606391, lay within its frozen interval 0.572160--0.630269 and was positive in all 11 networks. The contextual-information mean was $-0.015568$ versus reference interval 0.051178--0.109783, with four positive networks (one-sided $P=0.910$). Delayed geometry was $-0.000935$ versus interval 0.004511--0.006158, with no positive network ($P=1.000$). These temporal checks retain their specified lags and are reported separately from the five terminal-class endpoints. Lags are expressed in reference-cycle units. The result describes the tested composition and input schedule.

\subsection*{Controlled avoidance}

The controlled avoidance experiment used a virtual robot and a development-selected, frozen circuit. Five conditions comprised a matched sham and a two-by-two factorial of correct versus reversed sensory-to-action mapping and presence versus removal of downstream competition. Matched connections were redirected to inactive shadow destinations to preserve the graph and edge budget while removing the specified effect. Eight previously unseen networks were evaluated in three held-out worlds per condition, with 1,500 valid world ticks per run: 120 runs and 180,000 valid ticks. All 120 reported runs met the technical-validity checks.

Audits recorded static synaptic weights, no reward updates or collision-correction shortcut, at most one outstanding acknowledged input batch and no decision overshoot. These audits document the execution of the frozen circuit and the sensory--motor interface.

Collision burden was collision-active ticks divided by valid ticks. Clear-state and stop-command fractions used the same denominator. Risk clearance was the fraction of observed risk episodes ending within 25 world ticks. Undefined fractions remained undefined when there was no eligible denominator. For inference, world-level outcomes were averaged within each network before paired comparison. Pooled tick and episode totals were retained as descriptive quantities: the full-circuit risk-clearance rate was 99.82\% pooled and 99.81\% after network aggregation; the no-competition rates were 71.68\% and 68.80\%, respectively. A forward command was not interpreted as a measured distance travelled.

The primary full-circuit-minus-sham contrast used a two-sided exact sign-flip test over eight network differences, with Holm adjustment in the two-co-primary family. The companion predictive analysis is reported in the main Results and Supplementary Note 1, and the original correction family is retained. The reported descriptive interval was resampled at the network level. Gatekept collision contrasts used their specified correction family. Other action-state and behavioral probabilities are secondary and unadjusted unless explicitly stated. The laterality-by-competition interaction compared the competition effect under the two mappings. Closed-loop AUCs came from different experienced sensory trajectories; their comparison is not causal mediation under identical input, and a non-significant upstream contrast is not an equivalence test.

\subsection*{Companion predictive analysis}

A separate co-primary probe tested one frozen scalar score derived from the Atlas against new-circuit geometry. The probe comprised 20 operating points and eight independently generated networks, giving 160 runs. Within-network rank correlations summarized the relationship; the prespecified inference tested the positive direction. Mean correlation was $-0.213$, and every network had a negative correlation; the original positive-direction exact test gave $P=1.000$. A related seven-point embodied extension comprised 168 runs, with its association summarized across networks and a bootstrap interval. The score was fixed before evaluation and was not replaced or refitted after observing the outcomes. This analysis belongs to the same two-co-primary correction family as the full-circuit-versus-sham avoidance contrast. Supplementary Note 1 reports this predictor’s results.

\subsection*{Physical ANT robot and video}

The physical ANT robot uses approximately 1,400 synthetic neurons, four ultrasound sensors and a camera. Proximity observations and externally detected target evidence enter the neural circuit through the experiment interface. The circuit was configured before operation, and its ongoing dynamics select search, target approach and avoidance in an environment unmapped for the task. Software supplies sensing, encoding, the external visual detector, communication, actuator translation and safety handling. The neural control stage operates without a task-training dataset or backpropagation.

Supplementary Video 1 is a 57-second excerpt (source time 00:00--00:57) from the existing company demonstration recording. The external-camera panel has been spatially cropped to exclude the presenter and software interface, and the audio has been removed. The temporal order and playback rate within the excerpt are preserved; there are no newly inserted event cuts or generated frames. The source is an edited demonstration recording. Stills are from 4, 30 and 50 seconds of the excerpt. The resulting file is an H.264 MP4, supplied separately with its description.

The video provides a qualitative record of the real sensorimotor loop. It does not add trials to the virtual-robot avoidance cohort or supply an independent collision-success rate. The simulation's intervention comparisons and the physical demonstration therefore remain separately identified.

\subsection*{Embodied demonstrations}

The physical robotic-hand demonstration uses approximately 3,400 synthetic neurons and touch feedback to regulate grip. Feedback enters the neural inputs, the circuit produces control output, and actuation changes the next observation. The reported capability is closed-loop grip regulation on the physical hand.

The physical ANT demonstration and its recording are described under Physical ANT robot and video.

The drone demonstration uses approximately 1,500 synthetic neurons to control a virtual quadrotor in Webots. The reported behaviors are hovering and forward movement. Webots supplies the body and environment; the neural circuit supplies the rotor-control signals.

These demonstrations document sensing and action in different bodies. No trial-level success-rate estimate is reported for the hand, search robot or Webots drone. The 120-run controlled-avoidance cohort is separate (Methods, Controlled avoidance); its full-circuit condition was collision-free in 23 of 24 runs. Each demonstration has its own circuit, body, sensory interface and operating conditions.

\subsection*{Statistics and reproducibility}

Monte Carlo randomization, bootstrap uncertainty and exact paired inference served different purposes. Randomization tested the association against a defined null; bootstrap resampling estimated variability; exact sign-flip tests evaluated reproducibility across paired networks. The latter enumerated all $2^{11}$=2,048 or $2^{8}$=256 assignments, preserving the originally specified direction and statistic. Their exactness concerns the enumerated reference distribution under the sign-exchangeability assumption; it does not establish randomized treatment assignment or remove shared-stream and order confounds. One-sided and two-sided probabilities are labeled rather than interchanged. Bootstrap intervals resampled network pairs, not individual neurons, frames or lags; dense-time intervals are descriptive unless a simultaneous procedure is explicitly specified.

Holm adjustment\cite{holm1979} and Benjamini-Hochberg adjustment\cite{bh1995} use the specified endpoint families. Confirmatory endpoints, exploratory descriptors, selected engineering configurations and qualitative demonstrations are kept separate. Missing or unsupported observations are not recoded as zero, non-significance is not equivalence, and a decoded correlation is not itself a behavioral mechanism. Rounded display values are not used for inferential testing. Statistical decisions use the study-specific 0.05 threshold after the stated family adjustment where applicable. Exact sign-flip procedures use the paired network as the inferential unit and assume sign exchangeability under the null; they do not require a fitted normal distribution. Sample sizes were defined by the study designs described above. Formal a priori power calculations and blinded outcome assessment are not documented.

\subsection*{Figure preparation}

Figure 1 is a conceptual schematic of the OPU populations, their directed interaction and the resulting OOS. The population-level relation and illustrated OOS modes express the computational architecture.

Figures 3 and 5 reproduce the measured temporal-ablation and relation-memory plots. Figure 3 uses the functional labels phase mean, release mean and both averaged; its observations and axis scales are retained. Source Data supply the recruitment records, intrinsic-state observations, directed-relation summaries, network-level persistence values and the 121-cell persistence surface. Figure 5 expresses lags in target-cycle units with model-step equivalents given in its caption.

Figure 2 and Supplementary Figure 3 use operating-cell maps and frozen-profile network data. Figure 4 presents the three-terminal prospective-composition summaries; Table 2 adds the selected operating cells and their Atlas values. Figure 6 combines separate binary-stream and two-event summaries with a schematic associative pathway. Figure 7 uses paired avoidance and AUC observations and pooled behavioral fractions with a zero baseline. The numerical tables identify the aggregation level of each entry.

Supplementary Video 1 and its stills are extracted from the physical-robot recording as described above.

\subsection*{Manuscript preparation}

The author designed the experiments, performed the analyses, developed the argument and selected the references. Language-model assistance was used for language editing, drafting support and document formatting under the author’s direction. All experimental results come from the recorded studies, and the author takes full responsibility for the content.

\section*{Data availability}
Source Data accompany the article and contain the Figure 2 pair-level medians and operating maps, network-level frozen-profile and avoidance measurements, timing and composition summaries, and pooled behavioral values. The temporal-ablation tables provide recruitment records and intrinsic-state observations for Figure 3a,b, the plotted directed-relation means for Figure 3c, and paired network-level persistence for Figure 3d. Figure 5 is supported by network-level persistence curves and participation--persistence correlations and the 121-cell Theta-to-Delta persistence surface. The data dictionary identifies the aggregation level and figure mapping of each table. Supplementary Video 1 supplies the physical ANT demonstration excerpt. Seven-association session traces, remaining raw event records, configurations and study analyses may be requested from Alexander A. Wolf at research@artificialbrains.ai, identifying the experiment.

\section*{Code availability}

Figure-generation code and its input tables accompany the article. The public SDK documents the design and observation interface (\url{https://github.com/artificial-brains-inc/artificial_brains}). Requests for study-specific experiment-control, analysis and execution materials should be directed to the corresponding author, identifying the experiment. Reproduction uses its recorded configuration, input sequence, initial state, execution settings and analysis.

\section*{Acknowledgements}
This work was carried out at Artificial Brains.

\section*{Author contributions}
A.A.W. conceived the architecture, designed and performed the experiments, analyzed the data and wrote the manuscript.

\section*{Competing interests}
Alexander A. Wolf (Arturo Arguello) is the Lead Researcher of Artificial Brains, which develops the synthetic-neuron platform used in this study.

\begin{spacing}{1.1}

\end{spacing}

\clearpage
\setstretch{1.12}
\setlength{\tabcolsep}{4pt}

{\Large\bfseries Supplementary Information\par}
\vspace{8pt}
{\large\bfseries Oscillatory Processing Units as a computing substrate for biomimetic intelligence\par}
\vspace{8pt}
Alexander A. Wolf\\Lead Researcher, Artificial Brains, San Francisco, United States\\
ORCID: \href{https://orcid.org/0009-0005-4975-289X}{https://orcid.org/0009-0005-4975-289X}

\Needspace{6\baselineskip}
\section*{Supplementary Note 1 A separate predictive result}

The neural-dynamics and behavioral Atlas records measured relationships. A separate co-primary analysis tested whether one fixed scalar score derived from those relationships could predict the geometry of another circuit. The score did not predict the expected direction: the mean within-network rank correlation was $-0.213$, with all eight correlations negative (positive-direction $P=1.000$). Its related behavioral association was inconclusive (mean rank correlation 0.248; bootstrap interval $-0.0054$ to 0.5006; $P=0.0625$). The score was not refitted after those outcomes.

The scalar predictor was evaluated separately from the measured Atlas and the circuit operations. The avoidance contrast retains the original two-co-primary Holm correction. The study design and inference are described in the main Methods, Companion predictive analysis.

\clearpage
\section*{Supplementary Table 1 Study design and experimental units}

\small
\begin{longtable}{@{}p{0.23\linewidth}p{0.35\linewidth}p{0.34\linewidth}@{}}\toprule
\textbf{Study} & \textbf{Design and replication} & \textbf{Measured outcome}\\\midrule\endfirsthead
\textbf{Study} & \textbf{Design and replication} & \textbf{Measured outcome}\\\midrule\endhead
Factorial characterization & Four-class predictive study: 16 pairs; 121 cells each; 11 networks. Five-rhythm extension: 25 paired surfaces. & Participation and state descriptors with held-out validation; operating maps.\\[5pt]
Input-path portability & Two five-rhythm input paths; 3,025 matched cells, each across 11 networks. & Geometry prediction after realized-state matching.\\[5pt]
Prospective composition & Three terminals; ten branches; 11 new networks and held-out schedule. & Five locked response-class predictions; separate reference-interval and temporal checks.\\[5pt]
Temporal intervention & Three relations, four conditions and 11 networks; 99 altered runs plus 33 controls. & Loss of frozen-profile support in nine specified contrasts.\\[5pt]
Historical accessibility & 11 networks; one frozen 1,000-symbol stream; three receiving configurations. & Timing of historical readout with phase-preserving and future-label controls.\\[5pt]
Two-event memory & Eight networks; two conditions; 81 episodes per continuous run. & Context-transfer index and retained-past readout.\\[5pt]
Associative recall & Seven associations in one continuous network; separate two-object controls. & Terminal neural responses to partial cues.\\[5pt]
Arithmetic and Boolean operations & 152 technically valid runs; eight networks; two substrates. & Bounded addition, signed comparison and logical separation.\\[5pt]
Fuzzy operations & Five coarse input levels across three networks. & Complete output patterns and one downstream reuse.\\[5pt]
Decision studies & Selected timing composition with fresh confirmation sessions; separate single-trial panel. & History-conditioned mapping; support ordering and ties.\\[5pt]
Controlled avoidance & Five conditions; eight networks; three worlds per condition; 120 runs. & Collision burden, risk clearance and stopping.\\[5pt]
Embodied demonstrations & Physical hand and search robot; Webots drone. & Grip regulation, search, approach, avoidance, hover and forward movement.\\\bottomrule
\end{longtable}
\normalsize

\Needspace{6\baselineskip}
\section*{Supplementary Table 2 Frozen-profile probability loss}

Mean intact-minus-altered loss across 11 networks; all nine contrasts had 11 positive effects (nine-test Benjamini--Hochberg-adjusted $q=0.000488$). Condition labels match Supplementary Figure 3e.

\begin{tabular}{@{}llll@{}}\toprule
\textbf{Interaction} & \textbf{Phase mean} & \textbf{Release mean} & \textbf{Both averaged}\\\midrule
Beta-to-Theta & 0.9754 & 0.9957 & 0.9737\\
Gamma-to-Theta & 0.9790 & 0.9918 & 0.9919\\
Theta-to-Theta & 0.9735 & 0.9901 & 0.9371\\\bottomrule
\end{tabular}

\clearpage
\begin{figure}[p]
\centering
\includegraphics[width=\linewidth]{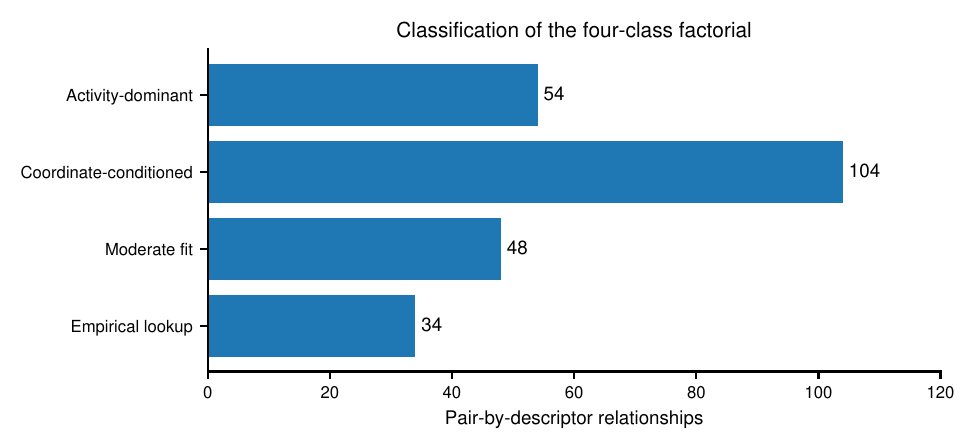}
\caption*{\textbf{Supplementary Figure 1. Firing participation and operating conditions answer different predictive questions.} Counts of the 240 directed-pair-by-descriptor relationships in the four-class mathematical Atlas. The activity-only model includes both source and target participation; the fuller model additionally retains their operating coordinates. Classification uses the conservative lower score from held-out source rows and target columns. Bars are category counts, not means or independent networks. There are no sampling error bars for these counts. Main Methods defines the classification and distinguishes this validation from held-out-network inference.}
\end{figure}

\clearpage
\begin{figure}[p]
\centering
\includegraphics[width=\linewidth]{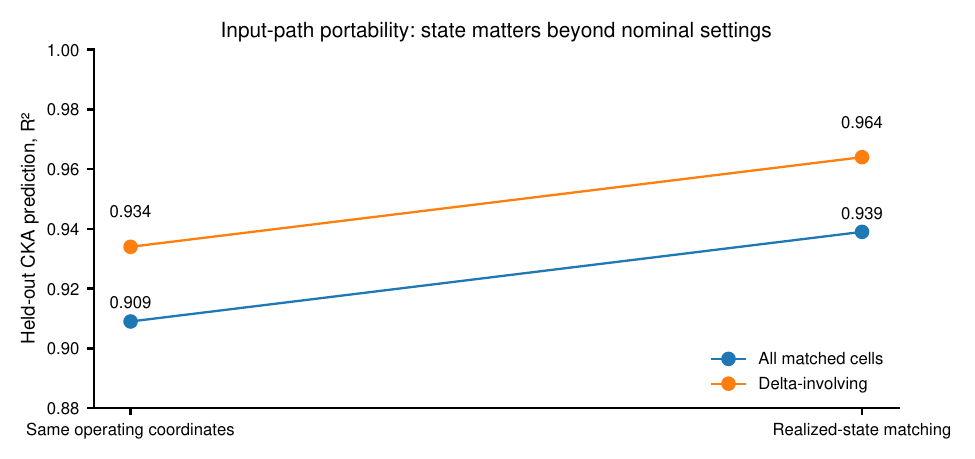}
\caption*{\textbf{Supplementary Figure 2. Realized-state matching improves transfer between input paths.} Held-out source--target CKA prediction across 3,025 matched operating cells in the two input-path architectures, with a separate summary for the 1,089 Delta-involving cells. The predictor changes from copying nominal coordinates to matching realized source and target participation. CKA is not used to fit that matching. Points are aggregate predictive scores, not independently replicated networks; joining them emphasizes the comparison and does not imply intermediate measurements. Main Methods describes the architecture comparison and the identifiability criterion.}
\end{figure}

\clearpage
\begin{figure}[p]
\centering
\includegraphics[width=\linewidth,height=.65\textheight,keepaspectratio]{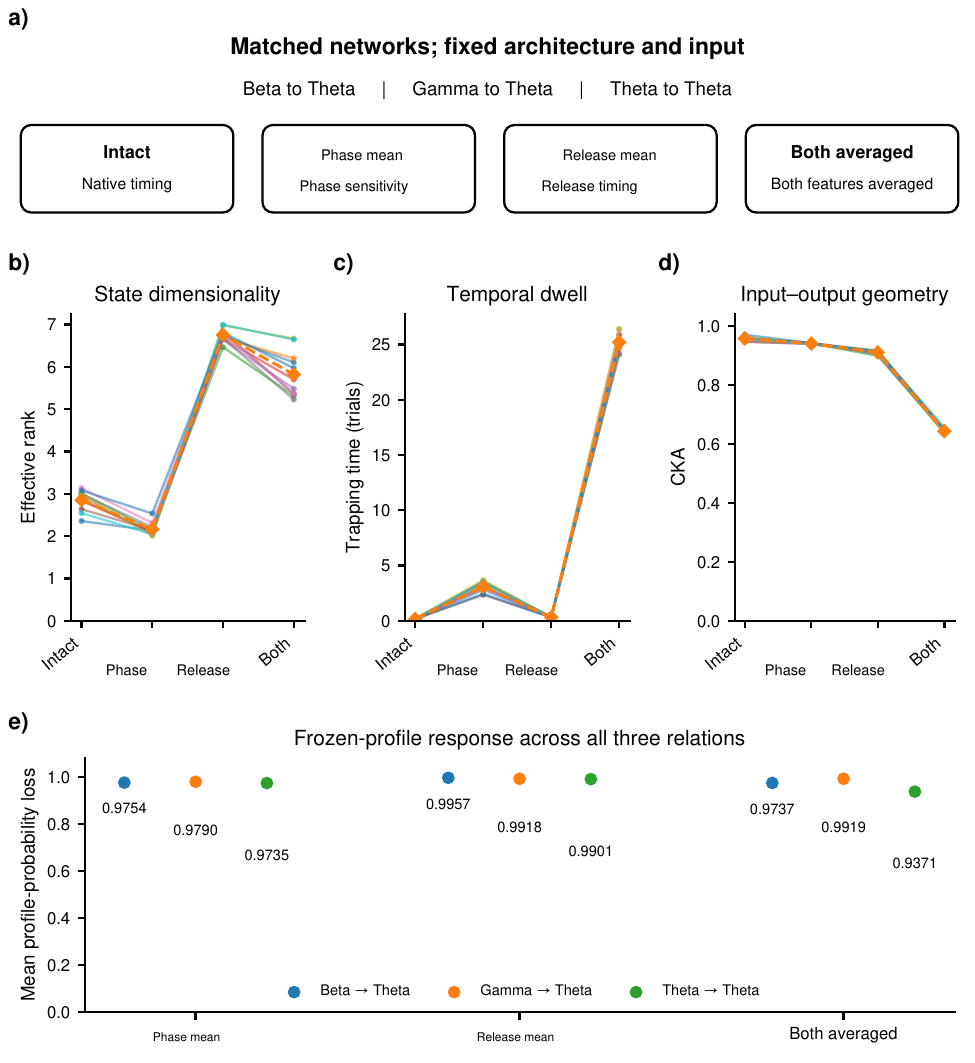}
\caption*{\textbf{Supplementary Figure 3. Temporal interventions change the receiving dynamics.} \textbf{a,} Experimental design: three source--Theta relationships, four temporal conditions and 11 matched networks. Phase sensitivity and periodic release were averaged separately or together. \textbf{b--d,} Network-level effective rank, trapping time and source--target CKA for the Theta-to-Theta relationship. Each network value averages all 121 operating cells. Thin lines connect the same network across conditions; diamonds and dashed lines show means. \textbf{e,} Mean loss of support for the frozen reference profile among its originally assigned cells, for all nine relationship-by-alteration comparisons. Every comparison was positive in all 11 networks (exact one-sided $P=0.000488$; nine-test Benjamini--Hochberg-adjusted $q=0.000488$). This profile-restricted endpoint is distinct from the full-surface descriptive measurements in b--d. The classifier was not refitted. Realized activity also changed, so the manipulation is not an activity-matched isolation of rhythm. Network observations and endpoint summaries are supplied separately.}
\end{figure}
\FloatBarrier

\end{document}